\documentclass{article}

\PassOptionsToPackage{numbers, compress}{natbib}
\usepackage[preprint]{neurips_2025}
\usepackage{float}
\usepackage{siunitx}

\usepackage{caption}
\usepackage[utf8]{inputenc} 
\usepackage[T1]{fontenc}    
\usepackage{hyperref}       
\usepackage{url}            
\usepackage{booktabs}       
\usepackage{amsfonts}       
\usepackage{nicefrac}       
\usepackage{microtype}      
\usepackage{xcolor}         
\usepackage{multirow}
\usepackage{makecell}
\usepackage{graphicx} %
\usepackage{wrapfig}
\usepackage{subcaption}
\usepackage{amsmath} 
\usepackage{makecell}
\usepackage{amsmath}
\usepackage{placeins} 
\title{GeoLocSFT: Efficient Visual Geolocation via Supervised Fine-Tuning of Multimodal Foundation Models} 

\author{
  Qiang Yi \\ 
  University of California, Berkeley\\ 
  \texttt{yiqiang20040822@berkeley.edu} \\ 
  \And 
  Lianlei Shan \\ 
  Tsinghua University \\ 
  \texttt{shanlianlei18@mails.ucas.edu.cn} \\ 
}

\begin{document}

\maketitle

\begin{abstract}

Accurately determining the geographic location where a single image was taken, visual geolocation, remains a formidable challenge due to the planet’s vastness and the deceptive similarity among distant locations. We introduce \textbf{GeoLocSFT}, a framework that demonstrates how targeted supervised fine-tuning (SFT) of a large multimodal foundation model (Gemma~3) using a small, high-quality dataset can yield highly competitive geolocation performance. GeoLocSFT is trained with only $\approx$2700 carefully selected image-GPS pairs from our geographically diverse MR600k dataset. Despite this limited data, our SFT-centric approach substantially improves over baseline models and achieves robust results on standard benchmarks such as Im2GPS-3k and YFCC-4k, as well as on our newly proposed and challenging MR40k benchmark, aimed specifically at sparsely populated regions. Further, we explore multi-candidate inference and aggregation strategies, but find that the core gains are already realized at the SFT stage. Our findings highlight the power of high-quality supervision and efficient SFT for planet-scale image geolocation, especially when compared to prior methods that require massive databases or complex pipelines. To foster further research, we publicly release the MR40k benchmark dataset.

\end{abstract}

\section{Introduction}
\label{sec:introduction}

\begin{figure}[t] 
  \centering
  \includegraphics[width=0.9\textwidth]{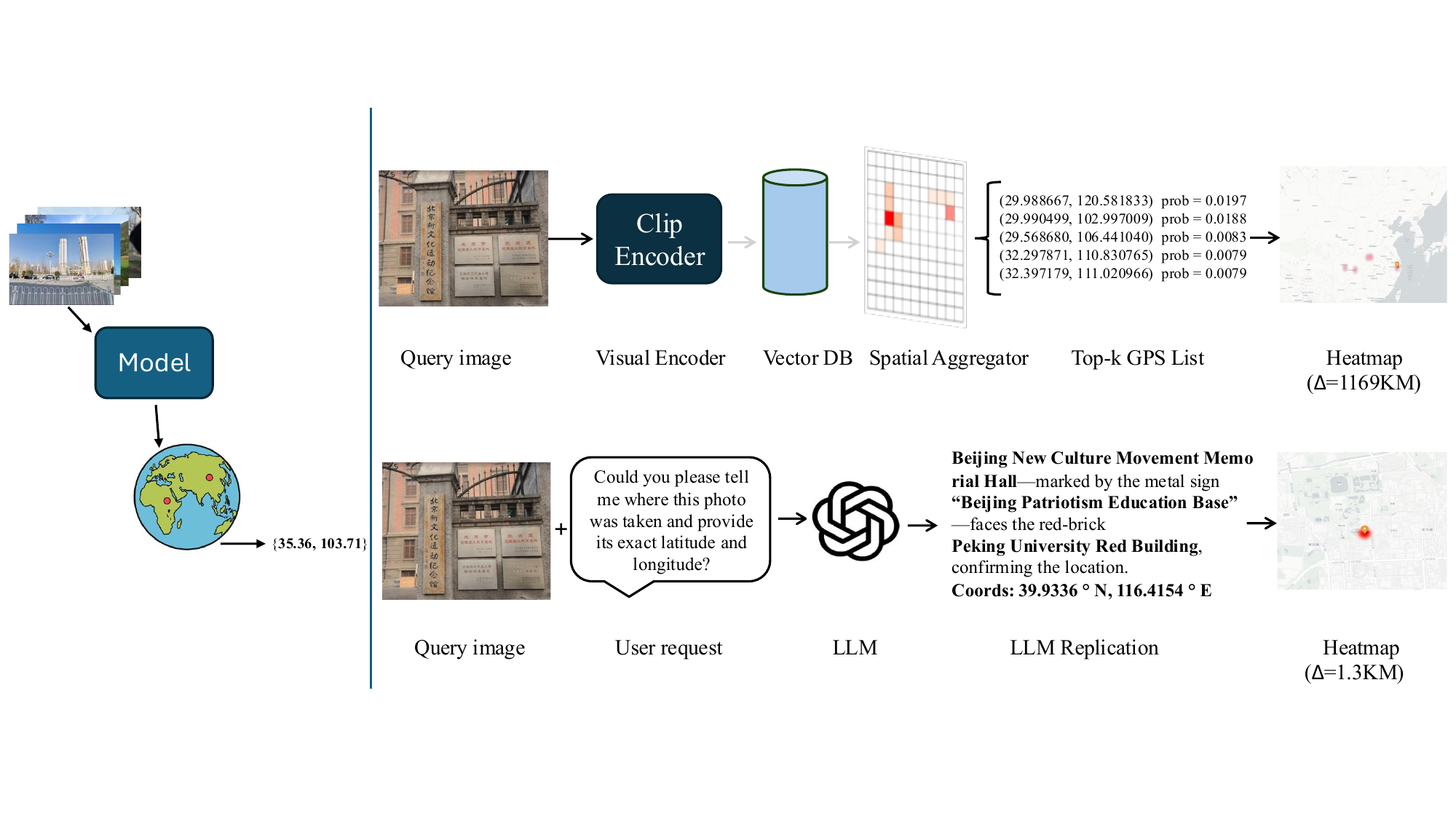} 
  \caption{Geolocation approach comparison. Traditional methods (Top) struggle with semantic details like text, leading to large errors (e.g., \SI{1169}{\kilo\metre}). Large Language models (Bottom) can use reasoning and world knowledge to interpret such fine-grained clues for significantly better accuracy (e.g., \SI{1.3}{\kilo\metre}).}
  \label{fig:intro_comparison}
\end{figure}

Estimating where a photograph was captured using only its pixels has long fascinated computer vision researchers. Accurate \emph{planet-scale} geolocation enables organizing photos by location, supports journalistic verification, and provides autonomous agents with a crucial fallback when GPS is unavailable. Recent breakthroughs in vision–language pre-training have made this task increasingly feasible: models such as CLIP \cite{radford2021learningtransferablevisualmodels}, trained on massive datasets of image–text pairs, already demonstrate notable zero-shot geolocation abilities by leveraging high-level semantic cues. However, solving global geolocation often requires more than just recognizing generic scene features—it demands the integration of diverse signals such as vegetation, architecture, signage, language, and subtle cultural markers into a coherent geographic understanding. Many images contain specific, highly informative details (e.g., text on signs, recognizable landmarks, or distinctive architectural elements) that call for fine-grained interpretation and reasoning. Traditional computer vision models, which frequently rely on global feature matching or coarse classification, often struggle with this level of nuanced understanding (see Figure~\ref{fig:intro_comparison}, Top). The ability to perform this kind of sophisticated, knowledge-driven reasoning from visual evidence (illustrated in Figure~\ref{fig:intro_comparison}, Bottom) is precisely what motivates the use of large foundation models for geolocation. As such, achieving accurate geolocation serves as a compelling benchmark for holistic scene reasoning in multimodal AI systems.

Despite notable advancements in visual geolocation, current methodologies still face significant limitations, which can be broadly categorized into two main paradigms. \emph{Early retrieval and classification approaches}, such as Im2GPS \cite{hays2008im2gps} (retrieval) and PlaNet \cite{Weyand_2016} (classification), rely heavily on large-scale geotagged image datasets, often comprising millions of examples. This dependence results in poor generalization, particularly in under-represented regions, and tends to promote memorization over genuine geographic reasoning. Moreover, these methods predominantly exploit global visual similarities, making them inadequate at capturing fine-grained semantic cues—such as text, distinctive local architecture, or subtle scene attributes—which are often crucial for accurate localization when overall scene appearance is ambiguous.
\emph{More recent large-scale frameworks}, including PIGEON \cite{haas2024pigeonpredictingimagegeolocations} and G3 \cite{jia2024g3effectiveadaptiveframework}, have achieved notable improvements in accuracy through sophisticated, multi-stage pipelines incorporating vision transformers, retrieval augmentation, clustering, and verification modules. However, such gains are accompanied by substantial drawbacks: these systems require enormous amounts of training data, incur significant computational costs, and are characterized by considerable implementation complexity, thus impeding reproducibility. Recent efforts to integrate explicit Chain-of-Thought (CoT) reasoning \cite{liu2024imagebasedgeolocationusinglarge} or Reinforcement Learning (RL) supervision have further compounded this complexity—often without delivering clear advantages over simpler, more direct approaches.
%

Consequently, there is an urgent need for frameworks that are efficient in both training and inference, require minimal data, and can generalize globally. Training and inference efficiency means models can leverage the inherent knowledge of modern foundation models, requiring limited computational resources and only a few inference steps. Low data requirements allow fine-grained visual understanding without vast databases or complex training schemes. Meanwhile, it is essential to address the ethical challenges of geolocalization to ensure responsible and safe use. Importantly, global generalization ensures that even with limited fine-tuning data, a model's general capabilities are preserved or improved. Our approach meets all three of these requirements.

In summary, we make several key contributions toward a more efficient and robust geolocation paradigm:
\begin{itemize}
\item We propose \textbf{GeoLocSFT}, a novel visual geolocation framework that leverages targeted Supervised Fine-Tuning (SFT) of a foundation model using a small, high-quality dataset. Our SFT-centric approach achieves highly competitive performance on standard benchmarks while relying on significantly less training data compared to previous state-of-the-art methods. Additionally, we explore an efficient inference strategy based on single-model sampling and provide an initial investigation into multi-candidate re-ranking (MCR).
\item We introduce \textbf{MR40k}, a new, challenging benchmark dataset for visual geolocation curated from Mapillary data. MR40k focuses on geographically informative imagery, providing a valuable resource for advancing geolocation research. Both MR40k and our code are publicly released to promote further research in this area.

\item Our findings demonstrate an efficient pathway to accurate planet-scale geolocation, underscoring the importance of data quality and effective SFT over sheer data volume or excessively complex training paradigms.
\end{itemize}

\section{Related Work}
\label{sec:related_work}

Our work builds upon prior research in visual geolocation and multimodal learning.

\textbf{Prior Geolocation Methods.} 
Early methods for image geolocation, like Im2GPS and PlaNet, used huge databases of geotagged images or classified images into geographic regions. These needed millions of images, struggled to generalize, and often just memorized rather than really reasoning about location. Recent advanced systems, such as PIGEON and G3, use complex, multi-step pipelines (with tools like Vision Transformers and retrieval augmentation) to get high accuracy, but they also require a lot of data and computational resources, making them hard to reproduce.
In contrast, our approach (GeoLocSFT) is much simpler and more efficient: we fine-tune a foundation model (Gemma 3) using supervised learning on a small, carefully chosen dataset (about 2,700 samples). This lets us predict coordinates directly without big reference databases or complicated pipelines at inference time.

\textbf{Reasoning and Foundation Models in Geolocation.} Leveraging the reasoning capabilities of Large Language Models (LLMs) is appealing. Some works explore explicit Chain-of-Thought (CoT) prompting \cite{wei2023chainofthoughtpromptingelicitsreasoning, liu2024imagebasedgeolocationusinglarge}, aiming for interpretable outputs. However, generating reliable reasoning chains and collecting necessary CoT supervision data globally remains challenging and costly, with performance benefits not always guaranteed. \textbf{Our Approach} leverages the foundation model's reasoning implicitly through SFT on geographically informative data, bypassing the need for explicit CoT supervision.

\textbf{Inference Strategies and Benchmarking.} While generating multiple outputs from a single model (cf. Self-Consistency \cite{wang2023selfconsistencyimproveschainthought}) can improve robustness, we also explore a Multi-Candidate Re-ranking (MCR) strategy. Our initial investigation into this strategy, evaluated in the ablation studies (Section~\ref{sec:experiments_ablations}, specifically the LLM Consensus results), aims to provide a task-specific aggregation method using geographic constraints and model confidence, differing from simple voting or traditional costly ensembles \cite{hansen1990neural}. To evaluate reasoning capabilities beyond standard benchmarks \cite{hays2008im2gps, Weyand_2016, Thomee_2016, clark2023werelookingatquery}, which may lack sufficient geographic cues, we introduce the MR40k dataset (Section~\ref{subsec:mr40k_benchmark_revised_v3}) derived from Mapillary \cite{neuhold2017mapillary}, focusing on informative images from sparsely populated regions.

\begin{figure}
    \centering
    \includegraphics[width=0.8\linewidth]{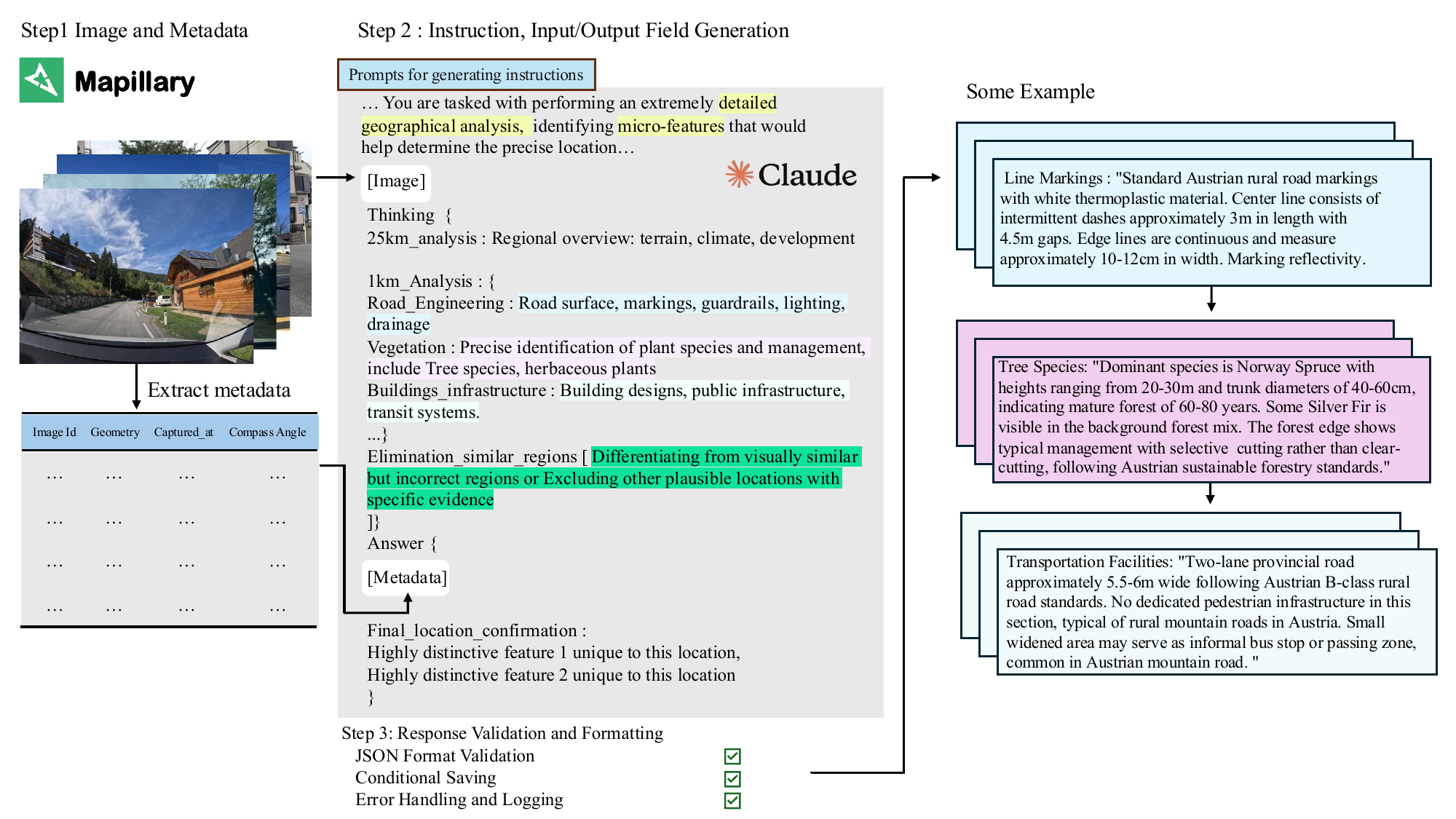}
    \caption{Generation pipeline for high-quality SFT "geo-captions". Step 1: Image/metadata extraction. Step 2: Claude 3.7 Sonnet (20k token budget) performs detailed geo-analysis via instructional prompting. Step 3: Structured JSON output validation. Examples (right) show fine-grained caption detail.}
    \label{fig:sft_data_building}
\end{figure} 

\section{Data Assets and Characteristics}
\label{sec:data_assets_and_characteristics_v3}

Our research leverages meticulously curated data assets for both training our GeoLocSFT framework and for comprehensive evaluation. This section details the foundational MR600k dataset, the generation process of our novel high-quality Supervised Fine-Tuning (SFT) data which forms a cornerstone of our contributions, and the MR40k benchmark designed for challenging geolocation in under-represented areas.

\subsection{MR600k: A Foundation of Diverse Street-Level Imagery}
\label{subsec:mr600k_foundation_revised_v3}
The MR600k dataset forms the bedrock of our data assets, comprising approximately 600,000 diverse, geolocated street-level images primarily sourced from Mapillary \cite{neuhold2017mapillary}. This repository provides a comprehensive visual representation of global environments, with each image accompanied by its GPS coordinates and associated metadata. From this foundational dataset, we meticulously select images for the generation of our high-quality supervised fine-tuning (SFT) data and for the construction of the MR40k benchmark, both of which are elaborated upon in the following discussions. 

\subsection{High-Quality SFT Data Generation: Eliciting Geographic Reasoning with LLMs}
\label{subsec:sft_data_generation_revised_v3}

A key part of this work is a new method for creating high-quality Supervised Fine-Tuning (SFT) data. We believe that to teach foundation models strong geographic reasoning, the training data must also show this type of reasoning. So, we built a pipeline that generates detailed, structured "geo-captions" for about 2,700 selected images from MR600k. As shown in Figure~\ref{fig:sft_data_building}, this process uses Claude 3.7 Sonnet to deeply analyze each image from multiple geographic perspectives.
The distinctiveness of our SFT data generation lies in the design of the instructional prompt and the analytical depth demanded from the LLM. Instead of simple object labeling or scene descriptions, our prompt compels Claude 3.7 Sonnet to act as an expert geographer. Key aspects that make our SFT data special include:

\textbf{Multi-Scale Contextual Analysis:} The LLM is required to analyze the image at both broad (e.g., 25km radius encompassing regional terrain, climate, development) and highly localized (e.g., 1km radius) scales. This ensures the geo-captions integrate macro-environmental context with micro-level visual details.

\textbf{Micro-Feature Identification and Interpretation:} The prompt explicitly tasks the LLM with identifying and interpreting a wide array of "micro-features" critical for precise geolocation. This process surpasses the model's standard object recognition capabilities by focusing on fine-grained details and their geographical significance, which are often crucial for distinguishing between visually similar locations. Examples include subtle variations in infrastructure, specific vegetation types, or unique architectural elements.

\begin{wrapfigure}{r}{0.5\textwidth}
  \centering
  \includegraphics[width=\linewidth]{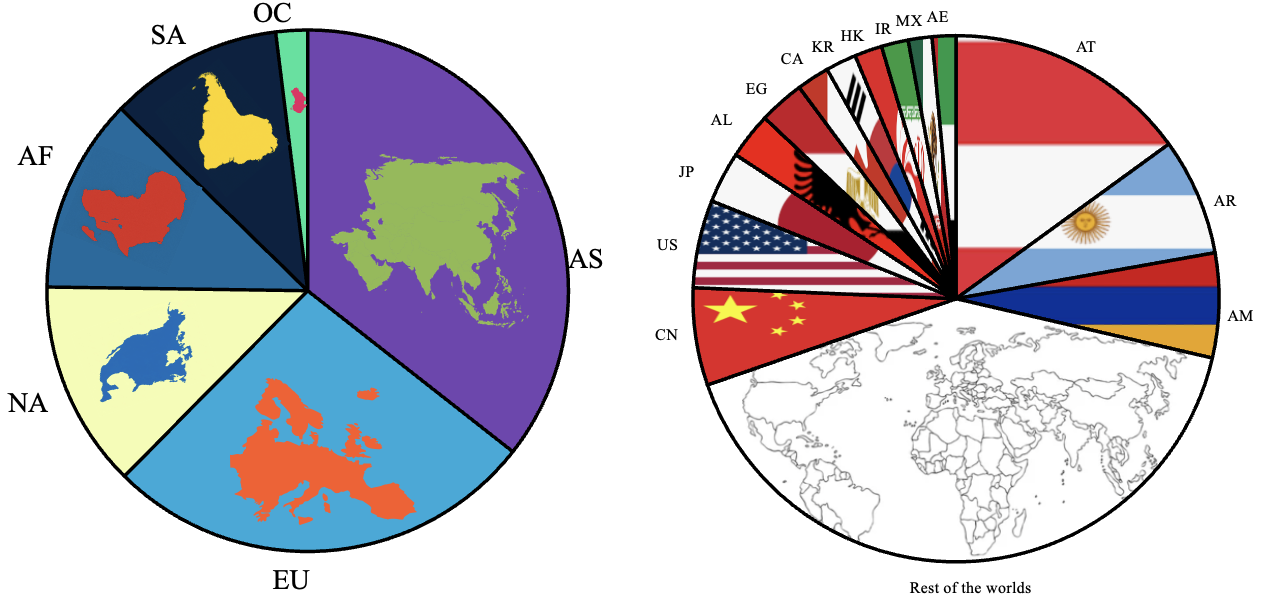} 
  \caption{Geographic distribution of the 2700 SFT
training images, detailing country/region specifics
and continental proportions.} 
  \label{fig:sft_distribution_wrap_experiment_sec_condensed} 
\end{wrapfigure}  

\textbf{Explicit Reasoning for Disambiguation:} A crucial component of our prompting strategy requires the LLM to perform comparative analysis by identifying potentially visually similar regions and then articulating specific, verifiable reasons (based on micro-features) why the input image cannot belong to those alternatives. This step directly elicits and encodes geographical disambiguation logic into the geo-captions, training the SFT model to differentiate between deceptively similar environments.

\begin{figure}[htbp] 
  \centering
  \includegraphics[width=0.9\textwidth]{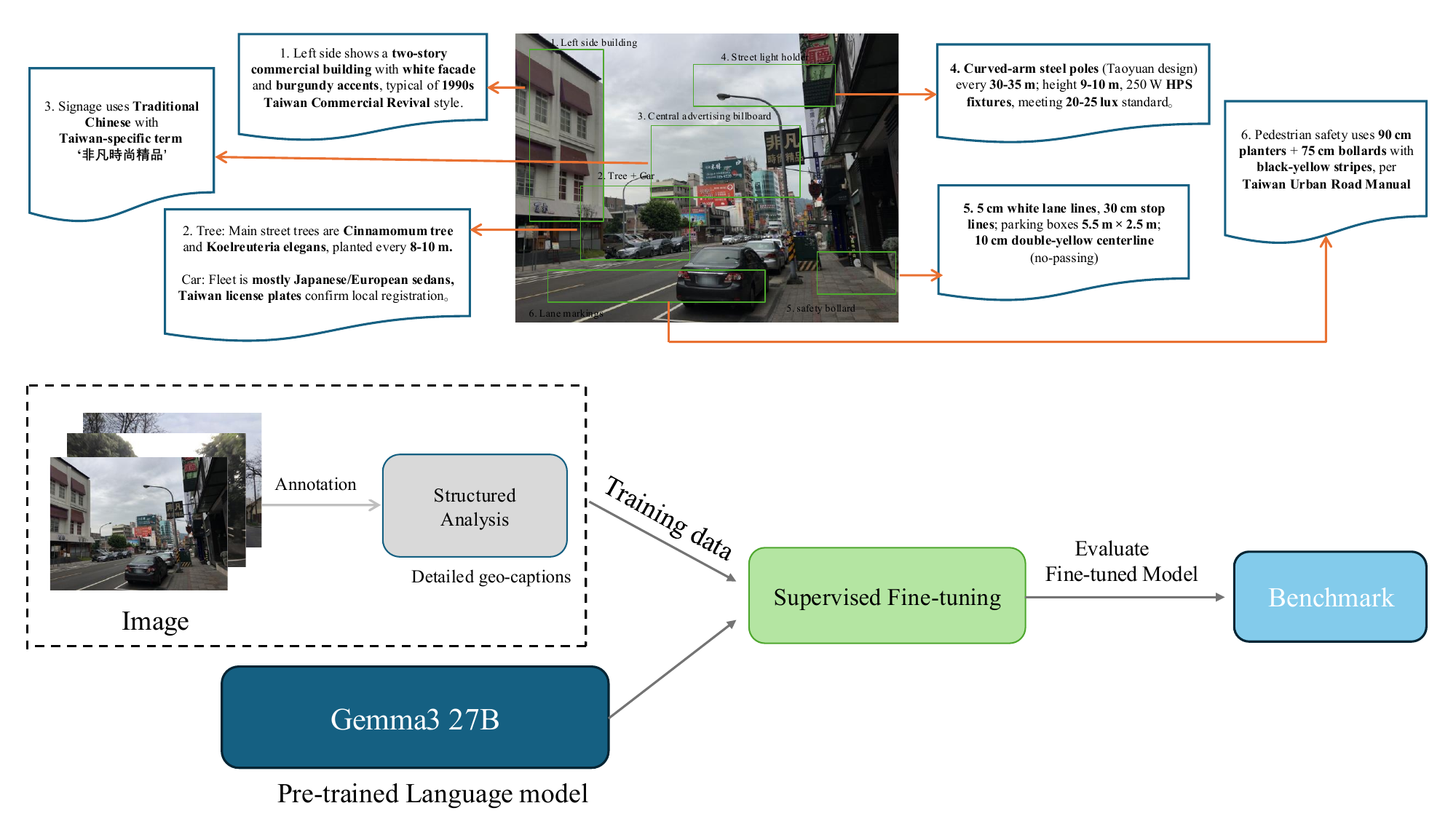} 
  \caption{Conceptual illustration of the Supervised Fine-tuning data preparation process.} 
  \label{fig:sft_process_experiment_sec_condensed} 
\end{figure} 

The LLM is prompted to generate a structured output, typically in JSON format, which methodically records these analytical components, aligning with the generation pipeline illustrated earlier. This process yields an SFT dataset of image-GPS pairs, where each image is annotated with a highly informative, multi-faceted "geo-caption." Far from being simple labels, these geo-captions embody analytical reasoning, integrate contextual knowledge, and articulate disambiguation strategies. It is this depth and quality of information, we argue, that renders them exceptionally effective as supervisory signals for fine-tuning. The substantial improvements observed in models trained with this data, a central finding of our experimental work, underscore the value of this approach to SFT data generation  

\subsection{MR40k: A Geolocation Benchmark for Sparsely Populated Regions}
\label{subsec:mr40k_benchmark_revised_v3}
\begin{wrapfigure}{r}{0.50\textwidth}
  \centering
  \includegraphics[width=\linewidth]{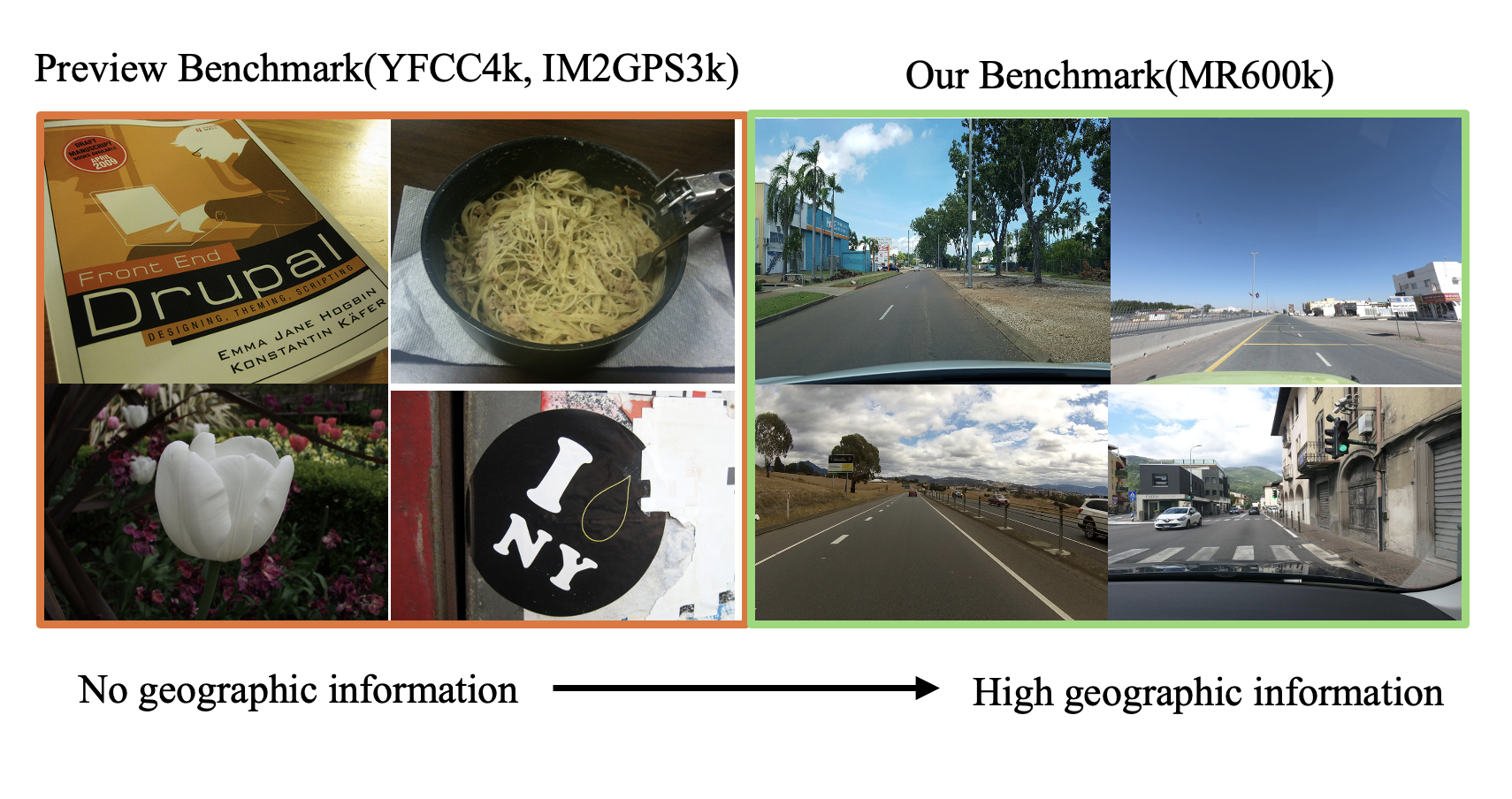} 
  \caption{Left: Examples from prior benchmarks often focusing on landmarks or urban areas. Right: MR40k examples showcasing street-level views from sparsely populated regions.}
  \label{fig:mr40k_benchmark_comparison_final} 
\end{wrapfigure}

To rigorously assess model generalization, particularly in non-urban and under-represented global environments, we introduce the \textbf{MR40k benchmark}. Existing benchmarks \cite{hays2008im2gps, Weyand_2016} often show a bias towards iconic landmarks or densely populated urban areas (Figure~\ref{fig:mr40k_benchmark_comparison_final}), which can limit the evaluation of a model's capacity for nuanced geographic reasoning in more diverse and challenging real-world settings.

Curated from our MR600k dataset, MR40k (40,000 images) was constructed using a targeted sampling strategy focused on \textbf{areas with an estimated population below 5,000 inhabitants}. This was achieved by filtering a global settlements database [cf. \cite{clark2023werelookingatquery, geonames_misc}] to identify such locations, sampling GPS coordinates therein, and retrieving the corresponding street-level images. Consequently, MR40k predominantly features globally distributed outdoor views from rural areas, small towns, and other less densely populated regions (Figure~\ref{fig:mr40k_benchmark_comparison_final}, Right examples), offering a distinct challenge from urban-centric datasets.

The explicit focus on low-population areas makes MR40k uniquely suited for evaluating model generalization and fine-grained reasoning beyond common scenarios. We will make MR40k publicly available\footnote{Link to dataset to be added upon publication or hosting.} to foster research into robust visual geolocation, especially in environments typically under-represented in existing resources. 

\section{Method}
\label{sec:method_balanced}

The GeoLocSFT framework introduces an efficient approach to visual geolocation, centered on the supervised fine-tuning (SFT) of large multimodal foundation models. This SFT strategy is uniquely powered by our high-quality "geo-caption" data, the generation of which—emphasizing detailed geographical analysis and reasoning—has been previously elucidated. In the following, we outline the core components of GeoLocSFT: the SFT process designed to adapt pre-trained models with this specialized dataset, and the resulting inference mechanism for geolocation prediction. 
\subsection{Overview}
\label{subsec:framework_overview_balanced}
The GeoLocSFT framework leverages the advanced reasoning of large multimodal foundation models like Gemma~3~\cite{gemmateam2025gemma3technicalreport} and Qwen2.5-VL~\cite{bai2025qwen25vltechnicalreport} for visual geolocation, primarily through a targeted supervised fine-tuning (SFT) process. 
This SFT utilizes our unique image--"geo-caption" pairs, rich in geographical analysis as previously detailed, to adapt the foundation model. 
During this offline phase, the model learns to map the visual and contextual logic within these geo-captions to precise coordinates. 
Subsequently, online inference is highly efficient, with predictions extracted from a single forward pass. 
While multi-candidate sampling is explored in our ablations, the core effectiveness of GeoLocSFT and its robust single-pass inference stem primarily from this potent SFT stage. 

\subsection{Supervised Fine-Tuning}
\label{subsec:sft_process_highly_streamlined}
We apply our Supervised Fine-Tuning (SFT) strategy, leveraging the rich “geo‑caption” data, to two distinct foundation models: the large-scale Gemma 3 27B-it \cite{gemmateam2025gemma3technicalreport} and the smaller Qwen2.5-VL-3B-Instruct \cite{bai2025qwen25vltechnicalreport}, to adapt them for the geolocation task. Effective SFT is the cornerstone of our approach.

\textbf{Fine-tuning Objective and Loss.} The primary objective of our fine-tuning is to train the model to predict the correct geographic coordinates $(\phi, \lambda)$ for a given input image $x$, by learning to generate the detailed "geo-caption" that includes these coordinates. We formulate this as a sequence generation problem. The model is trained to generate an output sequence $y = (y_1, y_2, ..., y_T)$ that mirrors the structure and content of the target geo-captions from our SFT dataset, including the textual reasoning process and the final coordinates enclosed in a specific format (e.g., \texttt{<answer> lat: ... lon: ... </answer>}).

We optimize the model parameters $\theta$ using the standard auto-regressive Cross-Entropy Loss over the curated SFT dataset $\mathcal{D}_{SFT}$:
\begin{equation}
\label{eq:sft_loss_method_final} 
L_{SFT}(\theta) = - \sum_{(x,y) \in \mathcal{D}_{SFT}} \sum_{t=1}^{|y|} \log P(y_t | y_{<t}, x; \theta),
\end{equation}
where $x$ is an input image, $y$ is its corresponding target geo-caption sequence (the stringified JSON output from Claude 3.7), and $P(y_t | y_{<t}, x; \theta)$ is the probability assigned by the model to the true token $y_t$ given the preceding tokens and the input image. 

\textbf{Fine-tuning Efficiency.} 
A key finding of our work is the remarkable efficiency with which effective model adaptation can be achieved. Across our fine-tuning strategies, merely \emph{one epoch} of training on our compact ($\approx$2700 samples) SFT dataset proves sufficient to elicit strong geolocation performance. This single-epoch approach significantly curtails computational costs compared to multi-epoch regimens, a conclusion supported by our subsequent ablation studies. For this SFT process, we employ the AdamW optimizer \cite{loshchilov2019decoupledweightdecayregularization} and utilize BF16 mixed-precision training. A comprehensive list of key hyperparameters is available in the appendix. 

\subsection{Inference}
\label{subsec:inference_balanced}
Geolocation prediction for a new query image is achieved via a single forward pass of the fine-tuned GeoLocSFT model. 
Given an input image, the model generates a textual output structured in the "geo-caption" format learned during supervised fine-tuning. 
The predicted geographic coordinates $(\phi_{pred}, \lambda_{pred})$ are then directly extracted from predefined \texttt{<answer>} tags within this generated text, serving as the final location estimate. 
While multi-candidate generation via sampling was explored, as detailed in our ablation studies, our primary findings highlight the strong performance achieved through this efficient single-pass inference, which captures the core benefits of our SFT strategy.

\section{Experiments}
\label{sec:experiments_corrected_layout}

\begin{figure}[htbp] 
    \centering
    \includegraphics[width=0.8\textwidth]{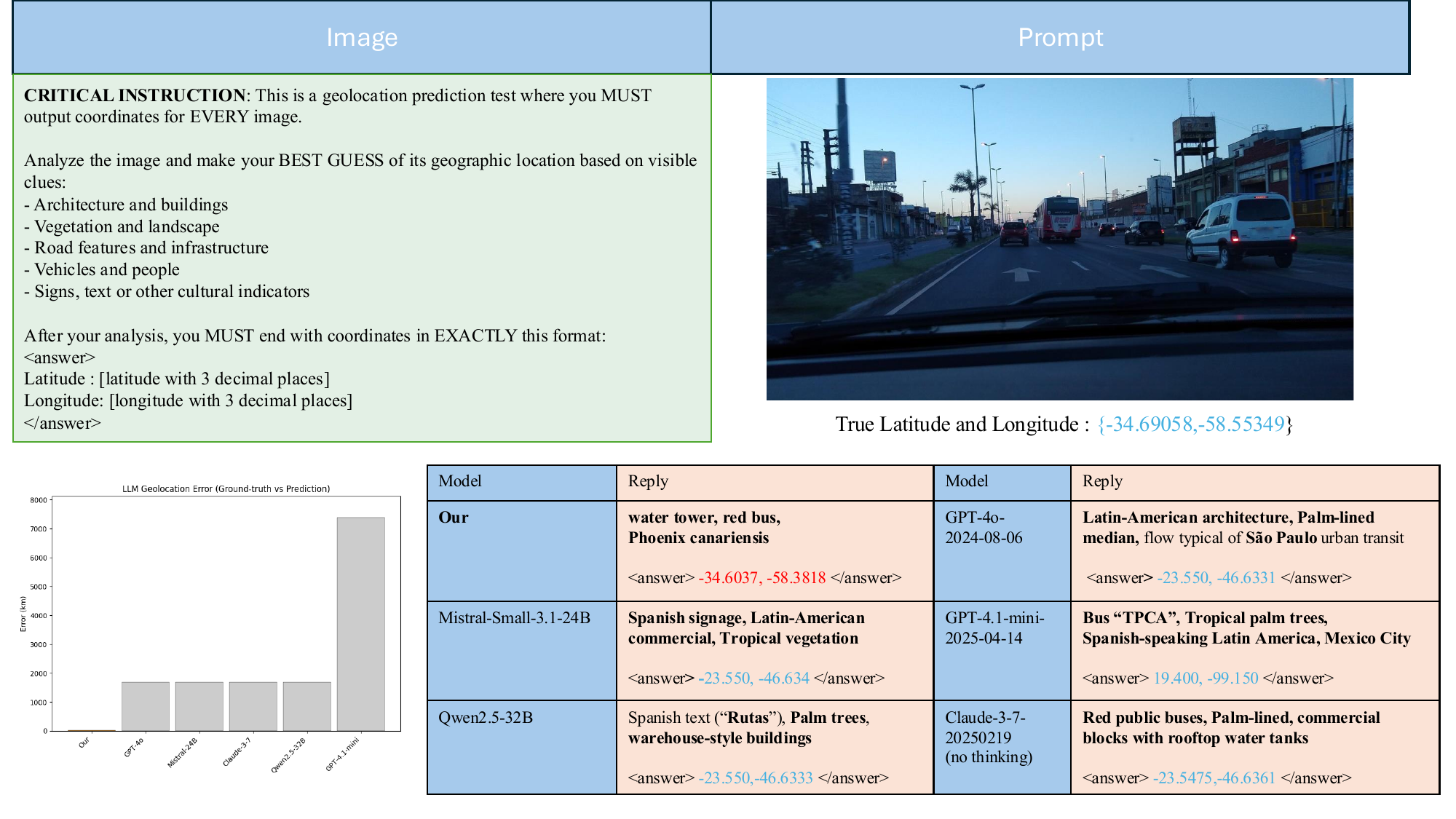} 
    \caption{Generic prompt and baseline model evaluation setup.} 
    \label{fig:baseline_prompt_example} 
\end{figure}

We evaluate GeoLocSFT by running experiments on standard visual geolocation benchmarks. We focus on testing how well our supervised fine-tuning strategy works, using high-quality "geo-caption" data as described earlier. We compare GeoLocSFT to other methods and perform ablation studies to understand the impact of different parts of our model. All experiments use standard datasets and Acc@R metrics. More details about the experimental setup can be found in Appendix~\ref{appendix:implementation_details}.

\subsection{Main Results and Analysis}
\label{subsec:results_corrected_layout}
We evaluated the impact of our SFT data by fine-tuning Gemma-3-27B-SFT (LoRA) and Qwen2.5 VL 3B SFT (complete fine-tuning). Their performance against respective baselines and other LMMs is detailed in Table~\ref{table:main_comparison_results_no_gpt4o}. Baseline models were prompted as illustrated in Figure~\ref{fig:baseline_prompt_example}.

\begin{wrapfigure}{r}{0.45\textwidth}
  \centering
  \includegraphics[width=\linewidth]{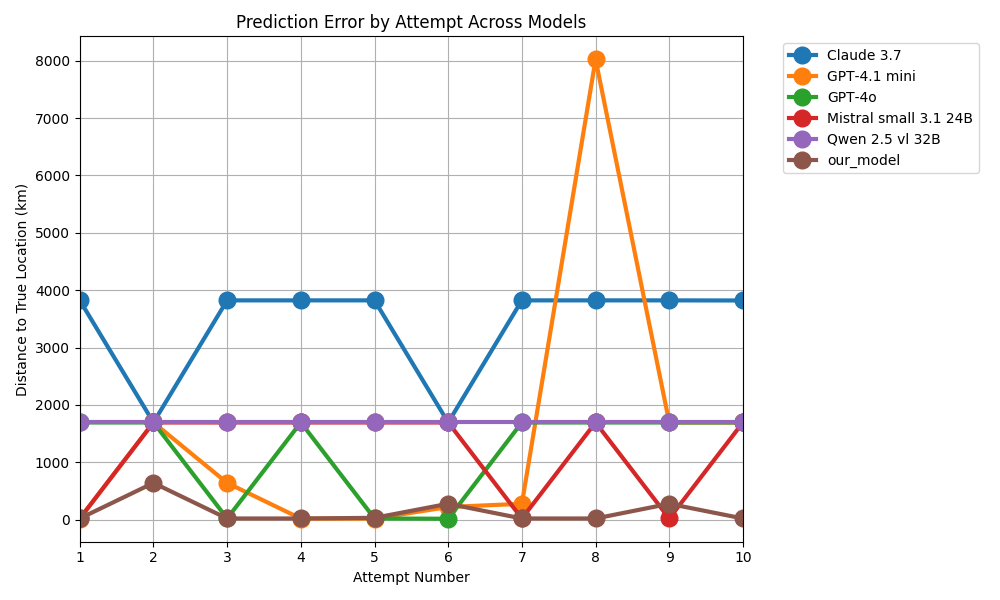} 
  \caption{Prediction error across $K=10$ attempts. GeoLocSFT models (e.g., Gemma 27B-SFT as 'our\_model') show lower variance.} 
  \label{fig:prediction_stability} 
\end{wrapfigure}

Our GeoLocSFT approach consistently enhances foundation model performance. Beyond aggregate accuracy, the nature of predictions is also refined by SFT. Figure~\ref{fig:prediction_stability} plots the prediction error across $K=10$ generation attempts for the same image using various models. While general-purpose models exhibit significant variance – note the large error spike for GPT-4.1 mini  and the fluctuating errors for others – our fine-tuned GeoLocSFT (Gemma 3 27B-SFT, labeled 'our\_model') demonstrates remarkably stable performance with consistently low prediction errors across all attempts. This stability suggests that SFT endows the model with a more robust and reliable geographical understanding for a given image, mitigating the volatility sometimes seen in zero-shot reasoning.

Interestingly, this stability in GeoLocSFT does not equate to deterministic output. Examining the sampled coordinates (further detailed in Appendix~\ref{appendix:mcr_analysis_examples}) reveals distinct behaviors: some large models, like Qwen 2.5 32B, produced identical outputs across all 10 attempts for this image (e.g., pinning to 23.550, -46.633), indicating potential mode collapse. Others, like GPT-4o and Claude 3.7 (from broader LMM evaluations), clustered around a very small number of ’safe’ locations. In contrast, GeoLocSFT (labeled 'our\_model' in the plot), while maintaining low error, generated a more diverse set of plausible coordinates primarily centered around the true location. This suggests SFT encourages a more nuanced exploration of the likely location space, potentially reflecting a better representation of uncertainty or finer-grained feature distinctions. This characteristic, generating focused yet diverse candidates, provides a stronger foundation for multi-candidate reasoning approaches.

\begin{table*}[htbp]
\centering
\caption{Geolocation accuracy (Acc@R, \%). GeoLocSFT models are fine-tuned with our high-quality "geo-caption" data. For each benchmark, we highlight the best general-purpose LMM and the best GeoLocSFT score. Full benchmark results appear in Table~\ref{table:full_main_comparison_results}.}
\label{table:main_comparison_results_no_gpt4o} 
\small
\setlength{\tabcolsep}{3.2pt} 
\begin{tabular}{l|l|ccccc}
\toprule
\multirow{2}{*}{Benchmark} & \multirow{2}{*}{Method} & \multicolumn{5}{c}{Distance (\% @ km)} \\
\cmidrule(lr){3-7}
& & 1 km & 25 km & 200 km & 750 km & 2500 km \\
\midrule

\multirow{8}{*}{\shortstack[l]{OSV5M\\(\cite{astruc2024openstreetview5mroadsglobalvisual})}} 
& Claude 3.7 Sonnet \cite{anthropic2024claude3}            & 0.85          & 3.41          & 13.91          & 40.21          & 61.97 \\
& Qwen2.5-VL-32B-Ins \cite{bai2025qwen25vltechnicalreport}  & 1.65          & 2.45          & 9.55           & 31.60          & 58.00 \\
& LLaVA 1.6-34b \cite{liu2023visualinstructiontuning}       & 0.00          & 0.05          & 1.40           & 10.80          & 32.00 \\
& Mistral-Small-3.1-24B-Ins \cite{mistralai2025mistralsmall31instructHF} & 0.05    & 0.60    & 5.25     & 23.95    & 48.55 \\
 
& Gemma-3-27b-it (Baseline) \cite{gemmateam2025gemma3technicalreport} & 1.74          & 2.94          & 11.76          & 35.70          & 61.61 \\
& GeoLocSFT (Gemma 3 27B-SFT)                             & \textbf{2.35} & \textbf{3.45} & \textbf{14.95} & \textbf{47.15} & \textbf{69.55} \\ 
& Qwen2.5-VL-3B-Ins (Baseline) \cite{bai2025qwen25vltechnicalreport}   & 0.40          & 0.60          & 4.25           & 20.15          & 41.50 \\
& GeoLocSFT (Qwen2.5-VL-3B-SFT)  & 0.55 & 0.85 & 8.05  & 31.10 & 55.20 \\ 
\midrule

\multirow{8}{*}{\shortstack[l]{IM2GPS3K\\(\cite{vo2017revisitingim2gpsdeeplearning})}} 
& Claude 3.7 Sonnet \cite{anthropic2024claude3}            & \textbf{17.96}         & \textbf{40.83}         & \textbf{50.66}          & 64.84          & 78.17 \\ 
& Qwen2.5-VL-32B-Ins \cite{bai2025qwen25vltechnicalreport}  & 8.90          & 31.90         & 47.10          & 63.80          & 79.95 \\
& LLaVA 1.6-34b \cite{liu2023visualinstructiontuning}       & 2.10          & 10.15         & 16.85          & 32.85          & 50.45 \\
& Mistral-Small-3.1-24B-Ins \cite{mistralai2025mistralsmall31instructHF} & 7.40    & 22.55   & 32.65    & 49.55    & 69.60 \\

& Gemma-3-27b-it (Baseline) \cite{gemmateam2025gemma3technicalreport} & 7.01          & 27.46         & 42.08          & 60.33          & 77.51 \\
& GeoLocSFT (Gemma 3 27B-SFT)                             & 8.80          & 32.70         & 47.20          & \textbf{65.80}          & \textbf{82.25} \\
& Qwen2.5-VL-3B-Ins (Baseline) \cite{bai2025qwen25vltechnicalreport}   & 2.75          & 19.25          & 34.40           & 55.00          & 68.15 \\
& GeoLocSFT (Qwen2.5-VL-3B-SFT)                            & 2.15          & 20.95          & 38.80           & 62.00          & 77.85 \\
\midrule

\multirow{8}{*}{MR40k (Ours)} 
& Claude 3.7 Sonnet \cite{anthropic2024claude3}  & 1.61 & \textbf{15.20}         & \textbf{39.57}          & 70.44           & 4.70 \\ 
& Qwen2.5-VL-32B-Ins \cite{bai2025qwen25vltechnicalreport}  & 0.65          & 7.80          & 30.25          & 57.85          & 74.20 \\
& LLaVA 1.6-34b \cite{liu2023visualinstructiontuning}       & 0.05          & 2.80          & 11.75          & 39.35          & 56.25 \\
& Mistral-Small-3.1-24B-Ins \cite{mistralai2025mistralsmall31instructHF} & 0.25    & 4.20    & 19.30    & 53.95    & 71.85 \\

& Gemma-3-27b-it (Baseline) \cite{gemmateam2025gemma3technicalreport} & 1.39          & 7.70          & 30.83          & 60.45          & 76.58 \\
& GeoLocSFT (Gemma 3 27B-SFT)                             & \textbf{1.70} & 12.76         & 37.55          & \textbf{70.85}          & \textbf{88.95} \\  
& Qwen2.5-VL-3B-Ins (Baseline) \cite{bai2025qwen25vltechnicalreport}   & 0.40          & 2.95          & 16.20           & 40.65          & 52.15 \\
& GeoLocSFT (Qwen2.5-VL-3B-SFT)                            & 0.60          & 4.45          & 20.70           & 50.50          & 68.90 \\
\bottomrule
\end{tabular}
\end{table*} 
 
\subsection{Austria Subset Performance}
\label{subsec:austria_subset_performance_corrected_final}

The capacity of our SFT data to instill deep regional understanding is strikingly evidenced by GeoLocSFT (Qwen2.5‑VL‑3B‑SFT)’s exceptional performance on an Austria‑specific test set (Table~\ref{table:austria_specific_results_final_label}). On this subset, our SFT‑enhanced 3B model outperformed strong general LLMs such as GPT‑4.1 across multiple distance thresholds, demonstrating that high‑quality, reasoning‑rich SFT data can empower even smaller models to develop expert‑level locale‑specific discernment by capturing subtle visual cues. GeoLocSFT (Gemma 3 27B‑SFT) likewise exhibited robust results on this subset (Table~\ref{table:austria_specific_results_final_label}).

\begin{table}[H] 
\centering
\footnotesize 
\setlength{\tabcolsep}{4.5pt}
\caption{Performance on the Austria Test Subset. Acc@R (\%).}
\label{table:austria_specific_results_final_label} 
\begin{tabular}{l|ccccc}
\toprule
Method & 1 km & 25 km & 200 km & 750 km & 2500 km \\
\midrule
Gemma-3-27b-it (Baseline) & 12.21 & 14.41 & 37.94 & 81.68 & 93.09 \\
OSV5M \cite{astruc2024openstreetview5mroadsglobalvisual} $\dagger$ & 0.00 & 7.71 & 39.54 & 77.08 & 93.69 \\
Qwen2.5-VL-32B-Ins & 11.21 & 15.12 & 48.45 & 84.79 & 91.69 \\
GPT-4.1-2025-04-14 & 10.91 & 20.62 & 58.96 & 86.99 & \textbf{96.00} \\
GeoLocSFT (Gemma 3 27B-SFT) & 12.91 & 17.02 & 50.05 & 87.19 & 95.60 \\
\textbf{GeoLocSFT (Qwen2.5-VL-3B-SFT)} & \textbf{13.81}  & \textbf{22.22} & \textbf{67.67} & \textbf{90.79}  & 95.10 \\
\bottomrule
\end{tabular}
\end{table} 

\subsection{Multi-Candidate Re-ranking Analysis}
\label{subsec:mcr_analysis_brief}

Beyond single-pass inference, we explored leveraging multiple prediction candidates (K=10) generated from GeoLocSFT (Gemma 3 27B-SFT) via sampling, coupled with a Multi-Candidate Re-ranking (MCR) strategy using LLM-based consensus. Table~\ref{table:mcr_sampling_aggregation_results_brief} compares single-prediction output (SFT-Single), the theoretical upper bound (Oracle Best K=10), and our MCR approach (LLM Consensus).

While Table~\ref{table:mcr_sampling_aggregation_results_brief} indicates a significant theoretical gap between SFT-Single and the Oracle Best, highlighting potential for improvement, our current MCR implementation yielded only marginal gains. The complexities and limitations of this MCR approach, including challenges in reliably assessing prediction confidence and handling ambiguous inputs, are further illustrated with qualitative examples in Appendix~\ref{appendix:mcr_analysis_examples}. These findings suggest that more sophisticated confidence estimation and fusion techniques are needed to fully realize the benefits of multi-candidate reasoning.

\begin{table}[H] 
\centering
\footnotesize 
\caption{Impact of Multi-Candidate Sampling (K=10) and MCR (LLM Consensus) on GeoLocSFT (Gemma 3 27B-SFT). Acc@R (\%).}
\label{table:mcr_sampling_aggregation_results_brief} 
\begin{tabular}{l|c|ccccc}
\toprule
Benchmark & Method & 1 km & 25 km & 200 km & 750 km & 2500 km \\
\midrule

\multirow{3}{*}{OSV5M}   & SFT-Single      &2.35 &3.45 &14.95 &47.15 &69.55  \\
                          & LLM Consensus  &2.73 &4.19   &16.68  & 49.12  &70.65  \\ 
                          & Oracle Best K=10 & \textbf{3.15} & \textbf{6.70} & \textbf{28.85} & \textbf{58.85} & \textbf{79.50} \\

\midrule
\multirow{3}{*}{Im2GPS3k} & SFT-Single      & 8.80 & 32.70 & 47.20 & 65.80 & 82.25 \\
                          & LLM Consensus   & 9.21 & 33.14 & 48.93 & 66.96 & 83.04 \\ 
                          & Oracle Best K=10 & \textbf{14.40} & \textbf{42.55} & \textbf{62.25} & \textbf{78.85} & \textbf{89.95} \\
\midrule

\multirow{3}{*}{MR40k}    & SFT-Single      & 1.70 & 12.76 & 37.55 & 70.85 & 88.95 \\
                          & LLM Consensus   & 1.76 & 15.11 & 39.30 & 73.49 & 89.99 \\ 
                          & Oracle Best K=10 & \textbf{2.60} & \textbf{24.15} & \textbf{61.35} & \textbf{86.50} & \textbf{94.00} \\
\bottomrule
\end{tabular}
\end{table} 

\subsection{Ablation Studies}
\label{subsec:ablation_studies_summary_with_key_table_ref}

To validate our design choices and component contributions, we performed several ablation studies. We first compared curated SFT data with a larger, weaker dataset for the Qwen2.5-VL-3B model, highlighting the importance of data quality (see Table~\ref{table:sft_quality_vs_quantity_ablation}). Additionally, we examined how different SFT data curation strategies (diverse vs. biased samples) and various hyperparameter settings, such as training epochs and LoRA configurations (mainly for the Gemma 3 27B model), affect performance.

The detailed methodologies, further results, and comprehensive discussions for all these ablation studies, including the data from Table~\ref{table:sft_quality_vs_quantity_ablation}, are provided in Appendix~\ref{sec:experiments_ablations} (Detailed Ablation Studies and Compared Method Details). The findings consistently support our main conclusions regarding the efficacy of our high-quality SFT data generation approach and the efficiency of our fine-tuning strategy.

\begin{table*}[htbp]
\centering
\caption{Ablation: Impact of SFT Data Quality vs. Quantity on Qwen2.5-VL-3B. "100k Unlabeled" uses weakly supervised data. Acc@R (\%).}
\label{table:sft_quality_vs_quantity_ablation} 
\footnotesize
\setlength{\tabcolsep}{1.0pt} 
\begin{tabular}{@{}l*{10}{c}@{}}
\toprule
\multirow{2}{*}{Methods (Qwen2.5-VL-3B)} & \multicolumn{5}{c}{GWS15K (\cite{clark2023werelookingatquery})} & \multicolumn{5}{c}{MR40K (Ours)} \\
\cmidrule(lr){2-6} \cmidrule(lr){7-11}
& \shortstack{Street\\1km} & \shortstack{City\\25km} & \shortstack{Region\\200km} & \shortstack{Country\\750km} & \shortstack{Continent\\2500km}
& \shortstack{Street\\1km} & \shortstack{City\\25km} & \shortstack{Region\\200km} & \shortstack{Country\\750km} & \shortstack{Continent\\2500km} \\
\midrule
Baseline & 0.05 & 3.95 & 28.70 & 61.30 & 79.00 & 0.40 & 2.95 & 16.20 & 40.65 & 52.15 \\
100k Unlabeled SFT & 0.05   &    3.90     &  28.55    &  61.05      & 78.80  & 0.38      & 2.88     &  16.05    &  40.40     & 51.95  \\
\textbf{GeoLocSFT (2700 Curated)} & \textbf{0.05} & \textbf{4.50} & \textbf{32.25} & \textbf{68.95} & \textbf{86.70} & \textbf{0.60} & \textbf{4.45} & \textbf{20.70} & \textbf{50.50} & \textbf{68.90} \\
\bottomrule
\end{tabular}
\end{table*} 

\section{Conclusion}
\label{sec:conclusion}

This paper introduced GeoLocSFT, a framework demonstrating that an efficient Supervised Fine-Tuning (SFT) strategy, using only $\approx$2700 meticulously curated image-"geo-caption" pairs, can substantially adapt large multimodal models (e.g., Gemma 3 27B-it \cite{gemmateam2025gemma3technicalreport} and Qwen2.5-VL-3B-Instruct \cite{bai2025qwen25vltechnicalreport}) for single-image visual geolocation with remarkable computational efficiency (typically one epoch). GeoLocSFT significantly outperforms respective baselines across diverse benchmarks, including our new MR40k dataset for sparsely populated regions, offering a resource-efficient alternative to methods reliant on massive datasets or complex pipelines. Our findings, supported by ablation studies, highlight the critical role of high-quality SFT data in achieving strong performance and even expert-level accuracy in specific regions (e.g., Austria subset). While multi-candidate sampling showed theoretical potential, current aggregation methods had limited success.

\newpage
\bibliographystyle{unsrtnat}
\bibliography{references}

\appendix
\clearpage 

\section{Qualitative Example: Nanjing Presidential Palace Geolocation Challenge}
\label{appendix:qualitative_nanjing}

This appendix presents a qualitative example illustrating the complexities of visual geolocation when fine-grained reasoning is required, contrasting the capabilities of foundation models with traditional approaches.

\begin{figure}[htbp]
    \centering
    \includegraphics[width=0.5\textwidth]{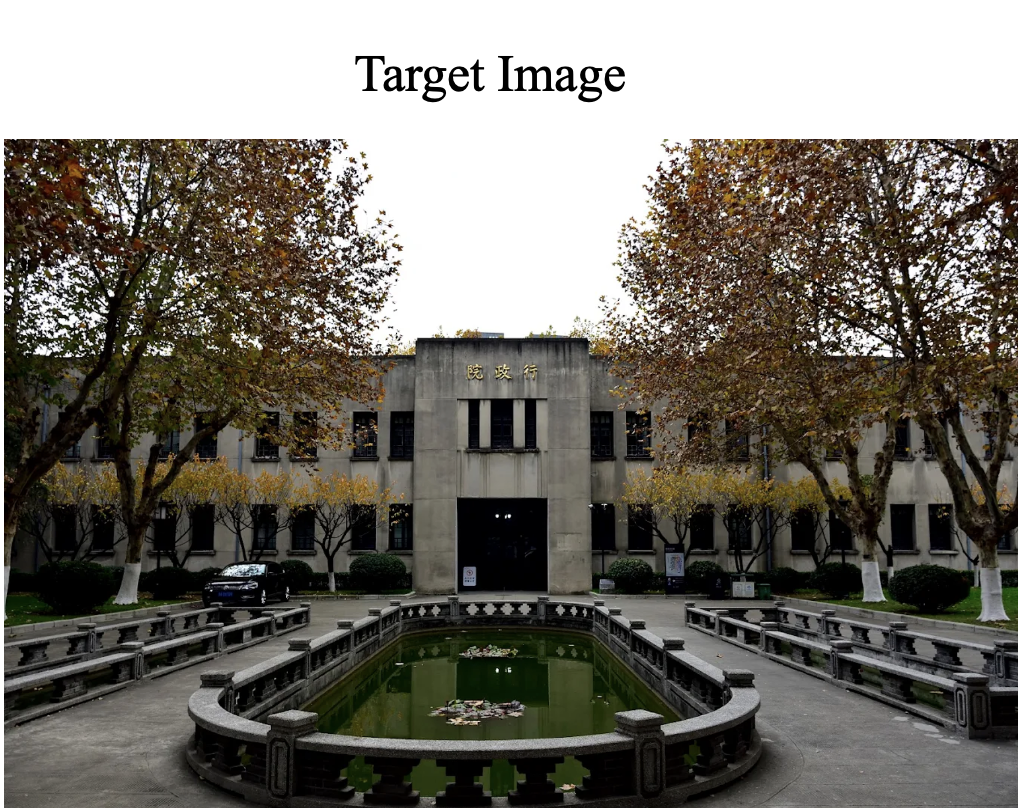} 
    \caption{Target Image: Former Executive Yuan at the Nanjing Presidential Palace. True Latitude and Longitude: {32.0456, 118.7922}.}
    \label{fig:appendix_nanjing_target}
\end{figure}

Figure~\ref{fig:appendix_nanjing_target} displays the target image, the Former Executive Yuan building within the Nanjing Presidential Palace complex in Nanjing, China. This location presents a challenge requiring recognition of specific architectural styles and potentially textual cues.

\begin{figure}[htbp]
    \centering
    \includegraphics[width=0.8\textwidth]{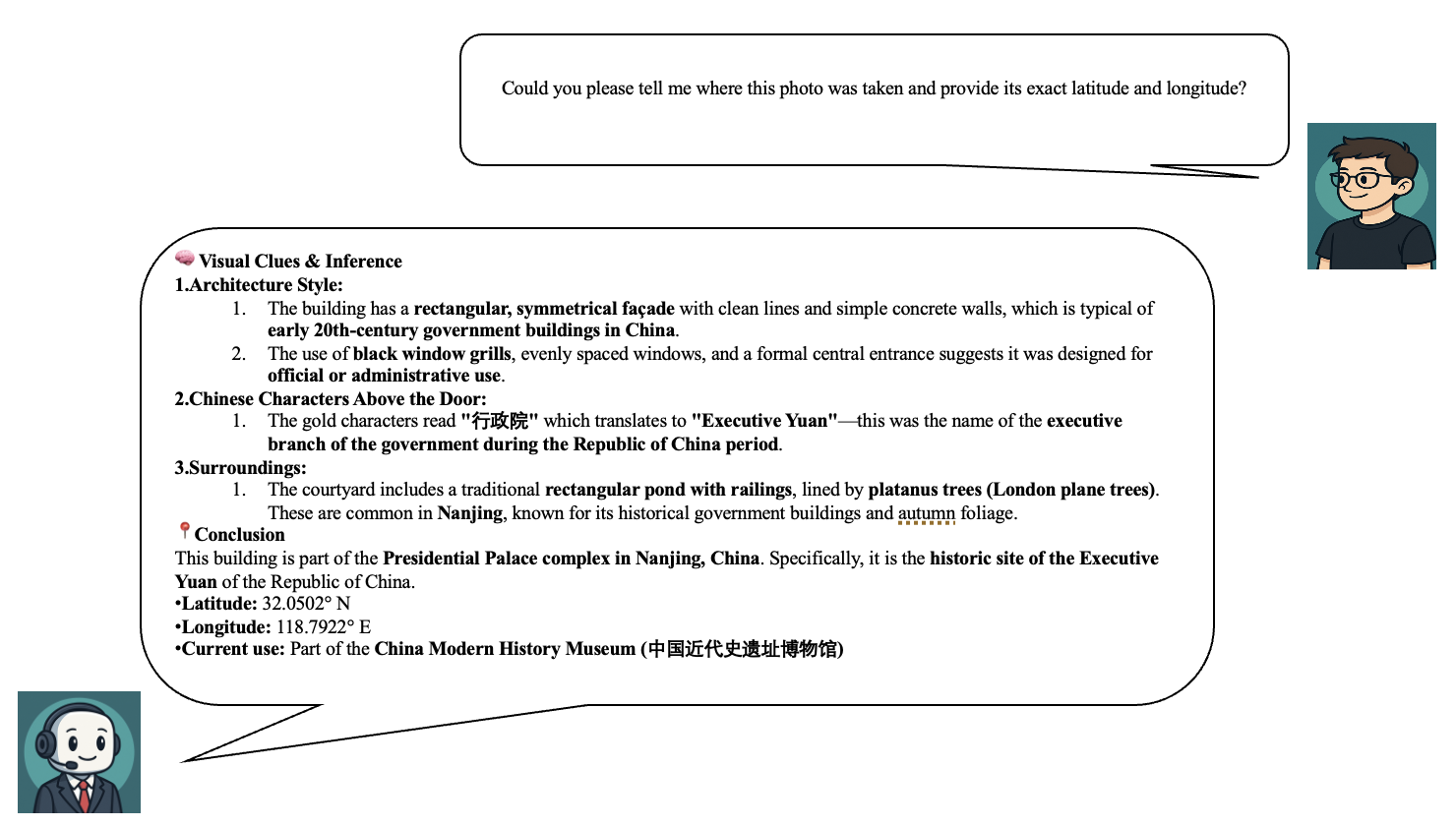} 
    \caption{Idealized analytical reasoning for the target in Figure~\ref{fig:appendix_nanjing_target}, simulating LLM-like analysis to pinpoint the location.} 
    \label{fig:appendix_nanjing_gpt_reasoning}
\end{figure}

An idealized analytical reasoning process for this target image is illustrated in Figure~\ref{fig:appendix_nanjing_gpt_reasoning}. This simulates how a powerful model might identify visual clues (e.g., specific architectural features indicative of a significant governmental or historical administrative building, surrounding foliage, and text on plaques or signs if legible) and integrate world knowledge to correctly deduce the location in Nanjing. 

\begin{figure}[!htbp]
    \centering
    \includegraphics[width=0.8\textwidth]{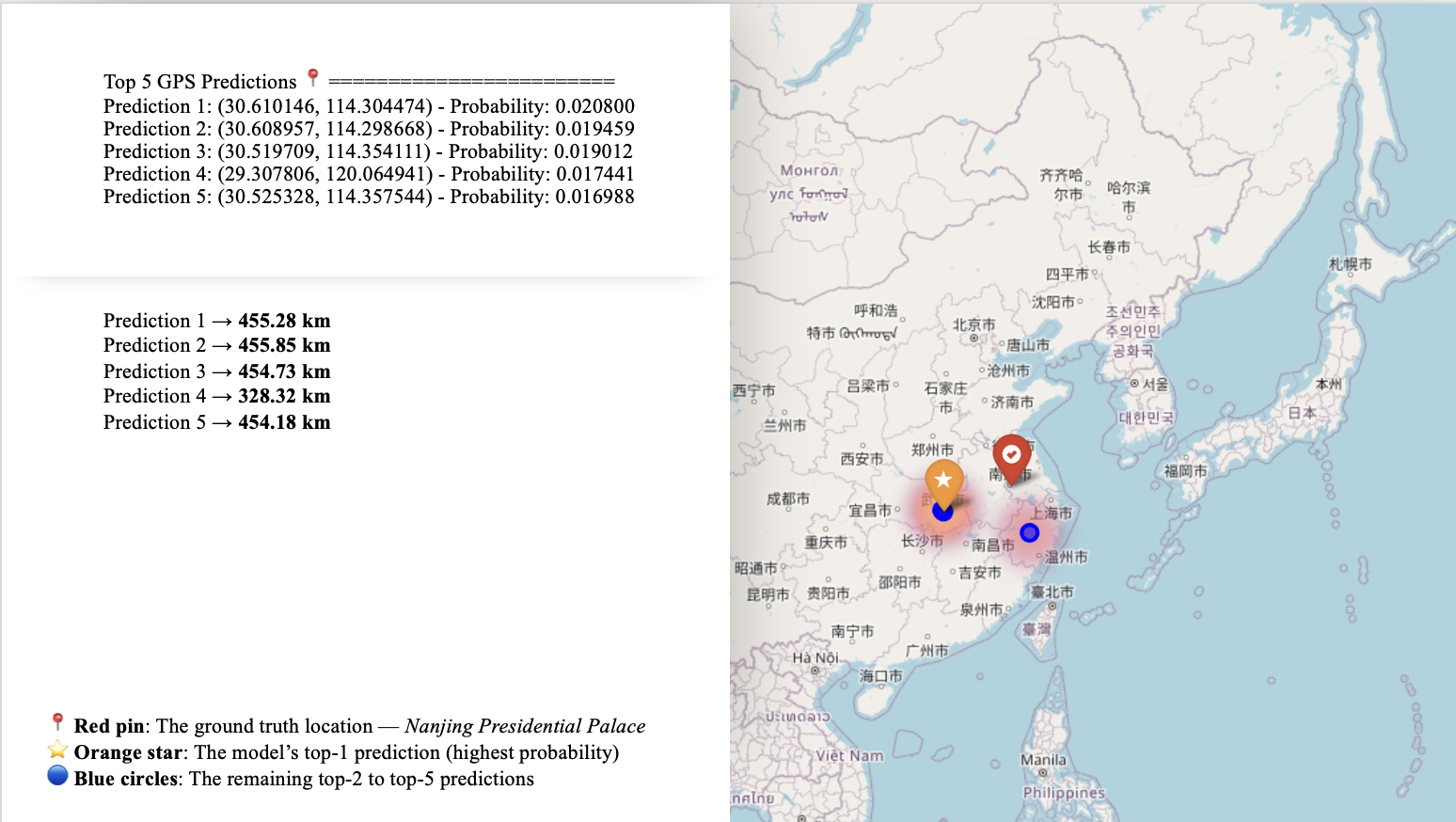}
    \caption{Actual Top 5 predictions from the GeoCLIP model~\cite{clark2023werelookingatquery} for the target image (Figure~\ref{fig:appendix_nanjing_target}), showing significant errors.}
    \label{fig:appendix_nanjing_geoclip_predictions}
\end{figure}
\FloatBarrier

In contrast, Figure~\ref{fig:appendix_nanjing_geoclip_predictions} shows the actual, erroneous Top 5 predictions from the traditional GeoCLIP model~\cite{clark2023werelookingatquery}. The map and prediction list clearly demonstrate large localization errors compared to the true location, highlighting the limitations of methods relying primarily on global visual feature matching when faced with scenes requiring nuanced contextual understanding and fine-grained detail interpretation. This example underscores the type of challenges our work aims to address.

\clearpage 
\section{Performance Comparison with General LLMs and Baseline Models}
\label{appendix:llm_baseline_comparison} 

\paragraph{Performance comparison with traditional geolocation methods.}
For a balanced evaluation, Table~\ref{table:main_results} contrasts our \emph{untuned} Gemma baselines and GeoLocSFT against several \emph{traditional} image-geolocation pipelines—mainly retrieval- or classification-based systems such as OSV5M, PIGEON, and G3.  
All legacy methods are run with the hyper-parameters reported in their original papers, whereas our generative models use the single generic prompt shown in Figure~\ref{fig:baseline_prompt_example}.

\begin{table}[H] 
\caption{Geolocation accuracy comparison across benchmarks and methods.}
\label{table:main_results}
\centering
\footnotesize
\setlength{\tabcolsep}{4.92pt}
\begin{tabular}{l|l|ccccc}
\toprule
\multirow{2}{*}{Benchmark} & \multirow{2}{*}{Method} & \multicolumn{5}{c}{Distance (\% @ km)} \\
\cmidrule{3-7}
& & 1km & 25km & 200km & 750km & 2500km \\
\midrule
\multirow{4}{*}{\shortstack[l]{OSV5M\\(\cite{astruc2024openstreetview5mroadsglobalvisual})}}
& OSV5M \cite{astruc2024openstreetview5mroadsglobalvisual} & 0.11 & \textbf{17.44} & \textbf{46.71} & 66.83 & \textbf{81.54} \\
& RFM \cite{dufour2024world80timestepsgenerative} & - & 5.4 & 44.2 & \textbf{76.2} & - \\
& Gemma3 27B \cite{gemmateam2025gemma3technicalreport} & 1.74 & 2.94 & 11.76 & 35.70 & 61.61 \\
& \textbf{GeoLocSFT (Ours)} & \textbf{2.35} & 3.45 & 14.95 & 47.15 & 69.55 \\
& $\Delta$ vs Best SOTA & +1.64 & -13.99 & -31.76 & -29.05 & -11.99 \\
\midrule
\multirow{5}{*}{\shortstack[l]{GWS15K\\(\cite{clark2023werelookingatquery})}}
& GeoCLIP \cite{clark2023werelookingatquery} & 0.6 & 3.1 & 16.9 & 45.7 & 74.1 \\
& PIGEON \cite{haas2024pigeonpredictingimagegeolocations} & \textbf{0.7} & 9.2 & 31.2 & 65.7 & 85.1 \\
& OSV5M \cite{astruc2024openstreetview5mroadsglobalvisual} & 0.11 & \textbf{17.44} & \textbf{38.25} & 66.83 & 81.54 \\
& Gemma3 27B \cite{gemmateam2025gemma3technicalreport} & 0.11 & 5.34 & 30.27 & 65.12 & 83.67 \\
& \textbf{GeoLocSFT (Ours)} & 0.2 & 6.3 & 33.50 & \textbf{69.65} & \textbf{87.55} \\
& $\Delta$ vs Best SOTA & -0.5 & -11.14 & -4.75 & +2.82 & +2.45 \\
\midrule
\multirow{5}{*}{\shortstack[l]{YFCC4k\\(\cite{vo2017revisitingim2gpsdeeplearning})}}
& OSV5M \cite{astruc2024openstreetview5mroadsglobalvisual} & 0.26 & 4.76 & 11.95 & 28.15 & 51.63 \\
& G3 \cite{jia2024g3effectiveadaptiveframework}& \textbf{23.99} & \textbf{35.89} & \textbf{46.98} & \textbf{64.26} & \textbf{78.15} \\
& PIGEON \cite{haas2024pigeonpredictingimagegeolocations} & 10.4 & 23.7 & 40.6 & 62.2 & 77.7 \\
& Gemma3 27B \cite{gemmateam2025gemma3technicalreport} & 3.46 & 13.89 & 26.72 & 46.34 & 61.53 \\
& \textbf{GeoLocSFT (Ours)} & 5.21 & 18.58 & 32.64 & 53.10 & 72.10 \\
& $\Delta$ vs Best SOTA & -18.78 & -17.31 & -14.34 & -11.16 & -6.05 \\
\midrule
\multirow{5}{*}{\shortstack[l]{IM2GPS3K\\(\cite{vo2017revisitingim2gpsdeeplearning})}}
& OSV5M \cite{astruc2024openstreetview5mroadsglobalvisual} & 0.83 & 13.28 & 25.33 & 43.84 & 64.63 \\
& G3 \cite{jia2024g3effectiveadaptiveframework} & \textbf{16.65} & \textbf{40.94} & \textbf{55.56} & 71.24 & 84.68 \\
& PIGEON \cite{haas2024pigeonpredictingimagegeolocations} & 11.3 & 36.7 & 53.8 & \textbf{72.4} & \textbf{85.3} \\
& Gemma3 27B \cite{gemmateam2025gemma3technicalreport} & 7.01 & 27.46 & 42.08 & 60.33 & 77.51 \\
& \textbf{GeoLocSFT (Ours)} & 8.80 & 32.70 & 47.20 & 65.80 & 82.25 \\
& $\Delta$ vs Best SOTA & -7.85 & -8.24 & -8.36 & -6.60 & -3.05 \\
\midrule
\multirow{4}{*}{MR40k (Ours)}
& OSV5M \cite{astruc2024openstreetview5mroadsglobalvisual} & 0.41 & \textbf{19.43} & \textbf{44.99} & \textbf{72.93} & 88.60 \\
& GeoCLIP \cite{clark2023werelookingatquery} & 0.38 & 8.89 & 37.41 & 71.35 & 88.94 \\
& Gemma3 27B \cite{gemmateam2025gemma3technicalreport} & 1.39 & 7.7 & 30.83 & 60.45 & 76.58 \\
& \textbf{GeoLocSFT (Ours)} & \textbf{1.70} & 12.76 & 37.55 & 70.85 & \textbf{88.95} \\
& $\Delta$ vs Best SOTA & +1.29 & -6.67 & -7.44 & -2.08 & +0.01 \\
\bottomrule
\end{tabular}
\end{table}

\newpage
\section{Complete Benchmark Results}
\label{sec:full_results}

Table~\ref{table:full_main_comparison_results} reports the end-to-end geolocation
accuracy (\textbf{Acc@R}) of all evaluated models on the five public
benchmarks used in this study—OSV5M, GWS15K, YFCC4k, IM2GPS3K—and on our
new MR40k dataset.  For each benchmark we list accuracy at five distance
thresholds (1~km, 25~km, 200~km, 750~km, 2\,500~km).  Bold numbers
highlight the strongest result within each benchmark row block,
separately for (i) general-purpose LMMs and (ii) our fine-tuned
GeoLocSFT variants.  These values are the source for all aggregate.

\begin{table*}[htbp]
\centering
\caption{presents the \emph{complete}
Acc@R scores for every model–benchmark pair evaluated in this work.
It supersedes the abridged results shown in the main text and serves as
the quantitative basis for all discussions in
Section~\ref{sec:experiments_corrected_layout}} 
\label{table:full_main_comparison_results} 
\small
\setlength{\tabcolsep}{3.2pt} 
\begin{tabular}{l|l|ccccc}
\toprule
\multirow{2}{*}{Benchmark} & \multirow{2}{*}{Method} & \multicolumn{5}{c}{Distance (\% @ km)} \\
\cmidrule(lr){3-7}
& & 1 km & 25 km & 200 km & 750 km & 2500 km \\
\midrule

\multirow{8}{*}{\shortstack[l]{OSV5M\\(\cite{astruc2024openstreetview5mroadsglobalvisual})}} 
& Claude 3.7 Sonnet \cite{anthropic2024claude3}            & 0.85          & 3.41          & 13.91          & 40.21          & 61.97 \\
& Qwen2.5-VL-32B-Ins \cite{bai2025qwen25vltechnicalreport}  & 1.65          & 2.45          & 9.55           & 31.60          & 58.00 \\
& LLaVA 1.6-34b \cite{liu2023visualinstructiontuning}       & 0.00          & 0.05          & 1.40           & 10.80          & 32.00 \\
& Mistral-Small-3.1-24B-Ins \cite{mistralai2025mistralsmall31instructHF} & 0.05    & 0.60    & 5.25     & 23.95    & 48.55 \\
 
& Gemma-3-27b-it (Baseline) \cite{gemmateam2025gemma3technicalreport} & 1.74          & 2.94          & 11.76          & 35.70          & 61.61 \\
& GeoLocSFT (Gemma 3 27B-SFT)                             & \textbf{2.35} & \textbf{3.45} & \textbf{14.95} & \textbf{47.15} & \textbf{69.55} \\ 
& Qwen2.5-VL-3B-Ins (Baseline) \cite{bai2025qwen25vltechnicalreport}   & 0.40          & 0.60          & 4.25           & 20.15          & 41.50 \\
& GeoLocSFT (Qwen2.5-VL-3B-SFT)  & 0.55 & 0.85 & 8.05  & 31.10 & 55.20 \\ 
\midrule

\multirow{8}{*}{\shortstack[l]{GWS15K\\(\cite{clark2023werelookingatquery})}} 
& Claude 3.7 Sonnet \cite{anthropic2024claude3}            & 0.19          & \textbf{11.09}         & \textbf{42.18}          & \textbf{72.89}          & \textbf{87.58} \\ 
& Qwen2.5-VL-32B-Ins \cite{bai2025qwen25vltechnicalreport}  & 0.00          & 5.65          & 27.25          & 62.70          & 82.75 \\
& LLaVA 1.6-34b \cite{liu2023visualinstructiontuning}       & 0.05          & 1.50          & 9.35           & 34.05          & 54.00 \\
& Mistral-Small-3.1-24B-Ins \cite{mistralai2025mistralsmall31instructHF} & 0.05    & 3.60    & 19.60    & 53.30    & 76.25 \\

& Gemma-3-27b-it (Baseline) \cite{gemmateam2025gemma3technicalreport} & 0.11          & 5.34          & 30.27          & 65.12          & 83.67 \\
& GeoLocSFT (Gemma 3 27B-SFT)                             & \textbf{0.20} & 6.30          & 33.50         & 69.65          & 87.55 \\
& Qwen2.5-VL-3B-Ins (Baseline) \cite{bai2025qwen25vltechnicalreport}   & 0.05          & 3.95          & 28.70           & 61.30          & 79.00 \\
& GeoLocSFT (Qwen2.5-VL-3B-SFT)                            & 0.05          & 4.50          & 32.25           & 68.95          & 86.70 \\
\midrule

\multirow{8}{*}{\shortstack[l]{YFCC4k\\(\cite{vo2017revisitingim2gpsdeeplearning})}} 
& Claude 3.7 Sonnet \cite{anthropic2024claude3}            & \textbf{10.80}         & \textbf{21.50}         & 32.01         & 48.48          & 60.89 \\ 
& Qwen2.5-VL-32B-Ins \cite{bai2025qwen25vltechnicalreport}  & 4.20          & 13.75         & 24.70          & 43.45          & 60.80 \\
& LLaVA 1.6-34b \cite{liu2023visualinstructiontuning}       & 0.85          & 4.80          & 8.80           & 22.00          & 36.70 \\
& Mistral-Small-3.1-24B-Ins \cite{mistralai2025mistralsmall31instructHF} & 3.55    & 10.70   & 18.55    & 36.75    & 57.65 \\

& Gemma-3-27b-it (Baseline) \cite{gemmateam2025gemma3technicalreport} & 3.46          & 13.89         & 26.72          & 46.34          & 61.53 \\
& GeoLocSFT (Gemma 3 27B-SFT)                             & 5.21          & 18.58         & \textbf{32.64}         & \textbf{53.10}          & \textbf{72.10} \\
& Qwen2.5-VL-3B-Ins (Baseline) \cite{bai2025qwen25vltechnicalreport}   & 0.85          & 6.60          & 14.30           & 27.65          & 38.30 \\
& GeoLocSFT (Qwen2.5-VL-3B-SFT)                            & 0.75          & 6.95          & 17.60           & 36.85          & 54.20 \\
\midrule

\multirow{8}{*}{\shortstack[l]{IM2GPS3K\\(\cite{vo2017revisitingim2gpsdeeplearning})}} 
& Claude 3.7 Sonnet \cite{anthropic2024claude3}            & \textbf{17.96}         & \textbf{40.83}         & \textbf{50.66}          & 64.84          & 78.17 \\ 
& Qwen2.5-VL-32B-Ins \cite{bai2025qwen25vltechnicalreport}  & 8.90          & 31.90         & 47.10          & 63.80          & 79.95 \\
& LLaVA 1.6-34b \cite{liu2023visualinstructiontuning}       & 2.10          & 10.15         & 16.85          & 32.85          & 50.45 \\
& Mistral-Small-3.1-24B-Ins \cite{mistralai2025mistralsmall31instructHF} & 7.40    & 22.55   & 32.65    & 49.55    & 69.60 \\

& Gemma-3-27b-it (Baseline) \cite{gemmateam2025gemma3technicalreport} & 7.01          & 27.46         & 42.08          & 60.33          & 77.51 \\
& GeoLocSFT (Gemma 3 27B-SFT)                             & 8.80          & 32.70         & 47.20          & \textbf{65.80}          & \textbf{82.25} \\
& Qwen2.5-VL-3B-Ins (Baseline) \cite{bai2025qwen25vltechnicalreport}   & 2.75          & 19.25          & 34.40           & 55.00          & 68.15 \\
& GeoLocSFT (Qwen2.5-VL-3B-SFT)                            & 2.15          & 20.95          & 38.80           & 62.00          & 77.85 \\
\midrule

\multirow{8}{*}{MR40k (Ours)} 
& Claude 3.7 Sonnet \cite{anthropic2024claude3}  & 1.61 & \textbf{15.20}         & \textbf{39.57}          & 70.44           & 4.70 \\ 
& Qwen2.5-VL-32B-Ins \cite{bai2025qwen25vltechnicalreport}  & 0.65          & 7.80          & 30.25          & 57.85          & 74.20 \\
& LLaVA 1.6-34b \cite{liu2023visualinstructiontuning}       & 0.05          & 2.80          & 11.75          & 39.35          & 56.25 \\
& Mistral-Small-3.1-24B-Ins \cite{mistralai2025mistralsmall31instructHF} & 0.25    & 4.20    & 19.30    & 53.95    & 71.85 \\

& Gemma-3-27b-it (Baseline) \cite{gemmateam2025gemma3technicalreport} & 1.39          & 7.70          & 30.83          & 60.45          & 76.58 \\
& GeoLocSFT (Gemma 3 27B-SFT)                             & \textbf{1.70} & 12.76         & 37.55          & \textbf{70.85}          & \textbf{88.95} \\  
& Qwen2.5-VL-3B-Ins (Baseline) \cite{bai2025qwen25vltechnicalreport}   & 0.40          & 2.95          & 16.20           & 40.65          & 52.15 \\
& GeoLocSFT (Qwen2.5-VL-3B-SFT)                            & 0.60          & 4.45          & 20.70           & 50.50          & 68.90 \\
\bottomrule
\end{tabular}
\end{table*}

\newpage
\section{Analysis of Multi-Candidate Re-ranking (MCR) Examples}
\label{appendix:mcr_analysis_examples}

This appendix illustrates the Multi-Candidate Re-ranking (MCR) mechanism with three representative examples (K=10 sampled candidates). Each figure combines the input image with its corresponding prediction list and true coordinates.

\begin{figure}[htbp]
  \centering
  \includegraphics[width=0.75\textwidth]{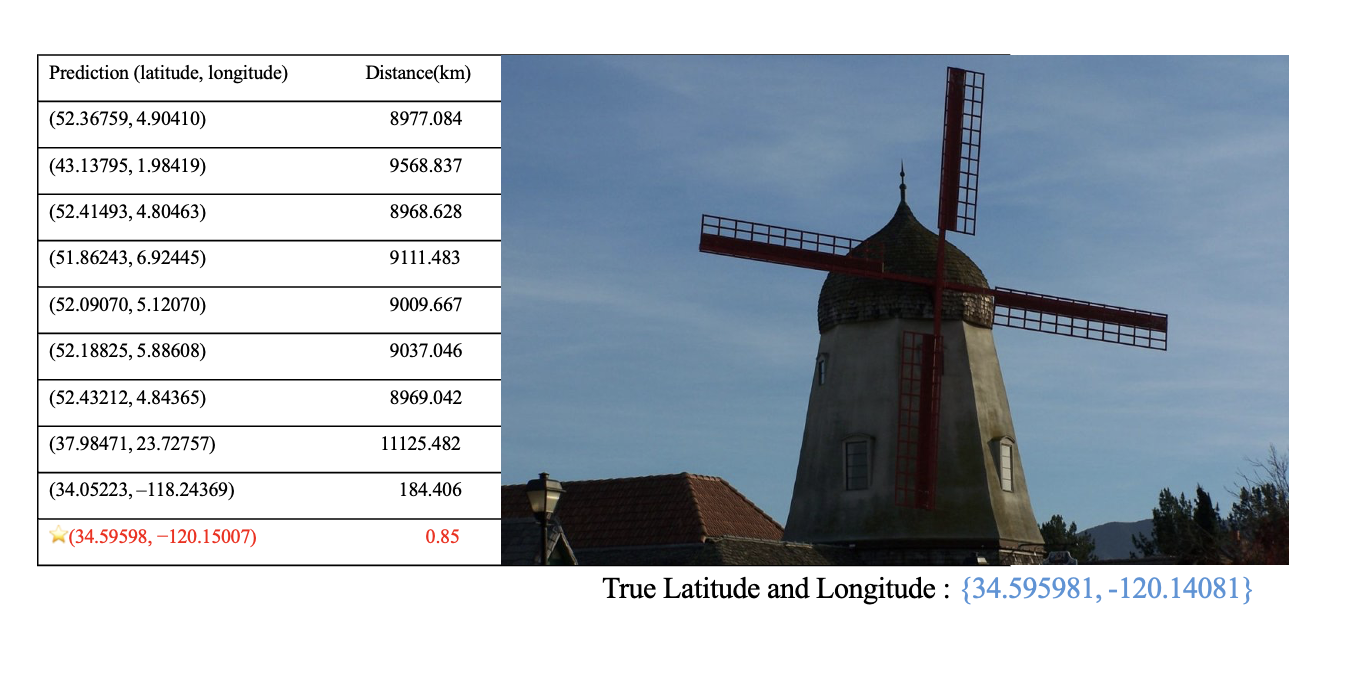} 
  \caption{MCR success case: Windmill image with prediction list. The red starred entry indicates the MCR-selected prediction. True Latitude and Longitude (Solvang, CA, USA): {34.595981, -120.14081}.}
  \label{fig:appendix_mcr_windmill_image}

\end{figure} 

\subsection{Example 1: Successful Identification of Strong Signal (Windmill)}
\label{subsec:appendix_c_windmill_success}
Figure~\ref{fig:appendix_mcr_windmill_image} shows an input image of a windmill (Solvang, CA, USA). While 9/10 predictions were highly inaccurate (>8900km error from the true location {34.595981, -120.14081}), one prediction (0.85km error, red star) was very accurate. MCR successfully selected this correct candidate.


\begin{figure}[htbp]
  \centering
  \includegraphics[width=0.75\textwidth]{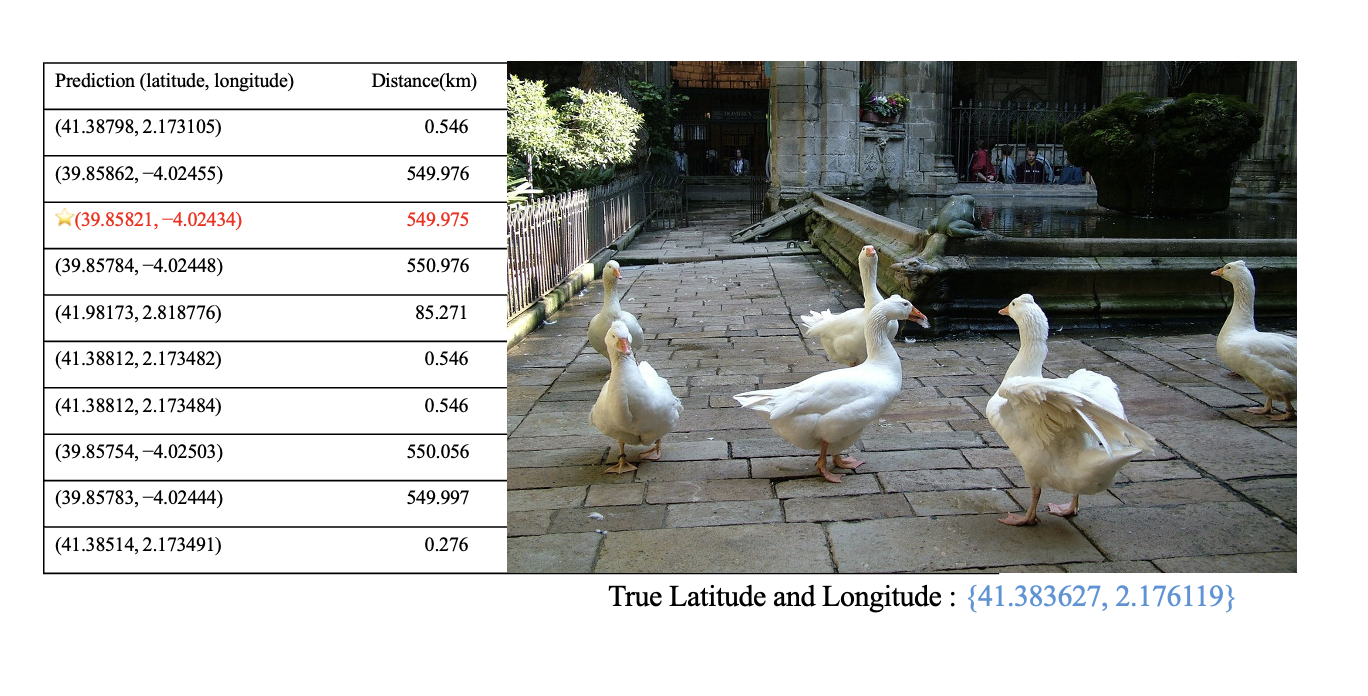} 
  \caption{MCR challenge: Landmark confusion (Geese image with prediction list). True location: Cloister of Barcelona Cathedral ({41.383627, 2.176119}). Red starred entry indicates MCR's selection pointing towards Toledo Cathedral.}
  \label{fig:appendix_mcr_geese_image}
\end{figure}

\subsection{Example 2: Challenge - Landmark Confusion (Geese in Courtyard)}
\label{subsec:appendix_c_geese_confusion}
Figure~\ref{fig:appendix_mcr_geese_image} presents an image of geese in a courtyard. The true location is the \textbf{Cloister of Barcelona Cathedral, Spain} ({41.383627, 2.176119}). Model predictions clustered around Barcelona (<1km error, 4 candidates) and erroneously near the \textbf{Toledo Cathedral, Spain} (~550km error, 5 candidates). MCR selected an incorrect prediction from the larger Toledo cluster (red star).

\begin{figure}[htbp]
  \centering
  \includegraphics[width=0.75\textwidth]{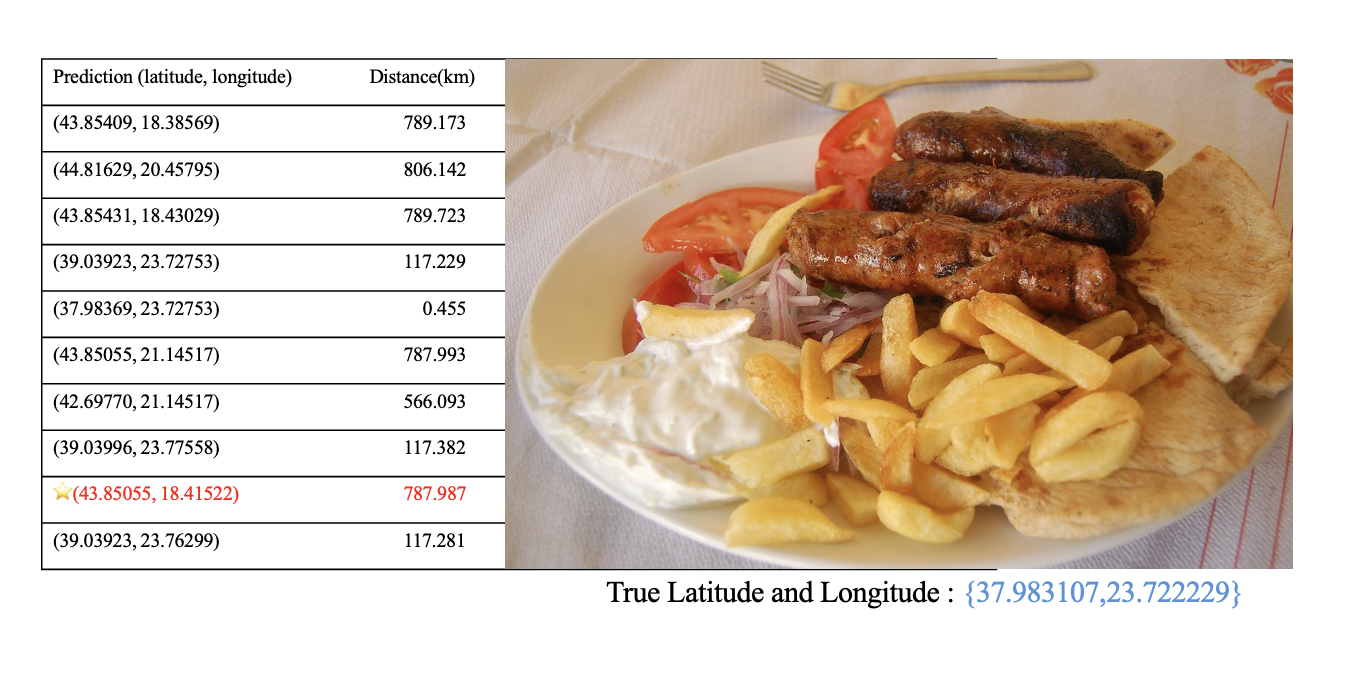} 
  \caption{MCR challenge: Input ambiguity (Food image with prediction list). True location: Athens, Greece ({37.983107, 23.722229}). Red starred entry indicates MCR's incorrect selection; a nearly correct prediction was not chosen.}
  \label{fig:appendix_mcr_food_image}
\end{figure} 

\subsection{Example 3: Challenge - Input Ambiguity / Neglected Correct Signal (Food)}
\label{subsec:appendix_c_food_ambiguity}
Figure~\ref{fig:appendix_mcr_food_image} shows a close-up food image; the true location is in central \textbf{Athens, Greece} ({37.983107, 23.722229}). One prediction was very close (0.455km error), but MCR selected a distant, incorrect prediction (788km error, red star). For ambiguous inputs, MCR may overlook isolated correct signals.

\subsection{Summary}
\label{subsec:appendix_c_mcr_summary}
These examples show MCR's utility with clear signals but also its limitations when faced with upstream model confusion or highly random outputs from ambiguous inputs.

\newpage
\section{Detailed Experimental Setup}
\label{appendix:implementation_details} 

This appendix provides comprehensive details of the experimental setup used in this work, including model configurations, supervised fine-tuning parameters, inference strategies, and hardware.

\subsection{Models and Baseline Setup}
\label{subsec:appendix_model_baseline_setup}
Our primary model, \textbf{GeoLocSFT}, is based on the Gemma 3 27B instruction-tuned model \cite{gemmateam2025gemma3technicalreport}.
The baseline models used for comparison are the unmodified, pre-trained Gemma 3 27B-it and Qwen2.5-vl-3B models. These baselines were evaluated by providing them with a generic instruction prompt for geolocation (as illustrated in Figure~\ref{fig:baseline_prompt_example} in the main text) and performing inference using temperature $T=1.0$ and top-p sampling with $p=0.95$. 

\begin{table}[htbp]
\centering
\caption{Supervised Fine-Tuning (SFT) parameters for GeoLocSFT.}
\label{tab:appendix_sft_parameters}
\begin{tabular}{@{}ll@{}} 
\toprule
Parameter & Value \\
\midrule
Base Model & Gemma 3 27B instruction-tuned (it) \cite{gemmateam2025gemma3technicalreport} \\
Fine-tuning Technique & Low-Rank Adaptation (LoRA) \cite{hu2021loralowrankadaptationlarge} \\
SFT Dataset & $\approx$2700 curated image-text pairs \\
& (from MR600k, see Sec.~\ref{subsec:mr600k_foundation_revised_v3} \& Sec.~\ref{subsec:sft_data_generation_revised_v3}) \\

\midrule
\multicolumn{2}{@{}l}{\textbf{LoRA Configuration}} \\ 
LoRA Rank ($r$) & 32 \\
LoRA Alpha ($\alpha$) & 32 \\
LoRA Dropout ($p$) & 0.05 \\
LoRA Target Modules & All linear layers of the Gemma model \\
\midrule
\multicolumn{2}{@{}l}{\textbf{Optimizer}} \\ 
Optimizer & AdamW \cite{loshchilov2019decoupledweightdecayregularization} \\
$\beta_1$ & 0.9 \\
$\beta_2$ & 0.999 \\
$\epsilon$ & $10^{-8}$ \\
\midrule
\multicolumn{2}{@{}l}{\textbf{Training Regimen}} \\ 
Learning Rate (Peak) & $8 \times 10^{-6}$ (8e-6) \\
LR Schedule & Cosine decay with 20 warmup steps \\
Training Epochs & 1 \\
Effective Batch Size & 16 \\
\textit{\small \hspace{0.5em} (Per-device: 2, Grad. Accum.: 1, GPUs: 8)} & \\ 
Loss Function & Cross-Entropy (auto-regressive over sequence) \\
\midrule
\multicolumn{2}{@{}l}{\textbf{Input/Output and Precision}} \\ 
Training Precision & BF16 mixed precision \\
Input Image Size & $895 \times 895$ pixels \\
Max. Sequence Length & 8096 tokens \\
\midrule
Training Framework & LLaMA Factory \cite{zheng2024llamafactoryunifiedefficientfinetuning} \\
\bottomrule
\end{tabular}
\end{table}

\subsection{Inference Strategies}
\label{subsec:appendix_inference_strategies}
\begin{itemize}
    \item \textbf{SFT-Single (for Main GeoLocSFT Results):} Our primary GeoLocSFT results reported in Table~\ref{table:main_results} were obtained via a single deterministic forward pass of the fine-tuned model. This involved greedy decoding (effectively temperature T=0 or very close to 0). Coordinates were extracted from the designated `<answer>...</answer>'' tags in the model's output. 
    \item \textbf{Sampling (for Analysis and MCR):} To generate $K=10$ candidates for the Oracle Best analysis, LLM Consensus aggregation, and MCR input, we used temperature $T=1.0$ and Top-p sampling with $p=0.95$. 
    \item \textbf{LLM Consensus Aggregation (for Analysis):} The 10 sampled textual outputs were processed using the fine-tuned GeoLocSFT (Gemma 3 27B) model, guided by a "Consensus-Geo" prompt designed to aggregate multiple predictions.
\end{itemize}

\subsection{Hardware and Training Time}
\label{subsec:appendix_hardware_training_time}
Supervised fine-tuning was performed on a cluster of 8 NVIDIA A100 (80GB) GPUs. The 1-epoch SFT process for the Gemma 3 27B model completed in approximately 50 minutes.

\newpage

\section{Detailed Ablation Studies and Compared Method Details}
\label{sec:experiments_ablations}

This appendix provides supplementary details for the ablation studies and lists the state-of-the-art (SOTA) methods compared against in the main paper.

\subsection{Compared State-of-the-Art (SOTA) Geolocation Methods}
\label{subsec:appendix_sota_methods_list}

Table~\ref{tab:appendix_sota_list} lists the SOTA geolocation methods compared against GeoLocSFT in Section~\ref{sec:experiments_corrected_layout}, along with brief descriptions. Results for these methods were sourced from their original publications or subsequent benchmark papers.

\begin{table}[!ht]
\centering
\caption{State-of-the-Art (SOTA) geolocation methods compared in this study.}
\label{tab:appendix_sota_list}
\begin{tabular}{@{}llp{0.6\textwidth}@{}} 
\toprule
Method & Citation & Brief Description \\
\midrule
PIGEON & \cite{haas2024pigeonpredictingimagegeolocations} & Utilizes Vision Transformers and semantic clustering, trained on large datasets. \\
G3 & \cite{jia2024g3effectiveadaptiveframework} & A multi-stage framework with retrieval-augmented generation from a knowledge base. \\
OSV5M & \cite{astruc2024openstreetview5mroadsglobalvisual} & A strong baseline model trained on the large OpenStreetView-5M dataset. \\
GeoCLIP & \cite{clark2023werelookingatquery} & Employs CLIP embeddings for image retrieval to determine location. \\
RFM & \cite{dufour2024world80timestepsgenerative} & A generative approach to global visual geolocation. \\
\bottomrule
\end{tabular}
\end{table}

\subsection{Impact of Training Epochs (1 vs 3)}
\label{subsec:appendix_ablation_epochs}

As discussed in Section~\ref{sec:experiments_ablations}, training for 3 epochs yielded only negligible or inconsistent improvements over a single epoch, while significantly increasing training time (approximately 3x).

\begin{table}[!ht]
\begin{tabular}{@{}p{0.48\textwidth}p{0.48\textwidth}@{}}
\begin{minipage}{\linewidth}
\scriptsize
\caption{Ablation Study: Impact of Training Epochs (1 vs 3). Gemma 3 27B-it fine-tuned on 2700 diverse samples (LR=2e-5, Rank=16). Acc@R shown in \%.}
\label{tab:appendix_epochs_full_data}
\setlength{\tabcolsep}{2pt}
\begin{tabular}{@{}llccccc@{}} 
\toprule
\multirow{2}{*}{Benchmark} & \multirow{2}{*}{Epochs} & \multicolumn{5}{c}{Distance (\% @ km)} \\ 
\cmidrule(lr){3-7} 
& & 1 km & 25 km & 200 km & 750 km & 2500 km \\ 
\midrule
\multirow{2}{*}{Im2GPS3k} & 1 Epoch & 8.70 & 32.45 & \textbf{47.00} & \textbf{65.25} & \textbf{82.15} \\
                          & 3 Epochs & \textbf{8.95} & \textbf{32.85} & 46.35 & 64.35 & 81.65 \\
\midrule
\multirow{2}{*}{GWS15k}   & 1 Epoch & \textbf{0.20} & 5.90 & \textbf{34.05} & \textbf{71.00} & 89.35 \\
                          & 3 Epochs & 0.10          & \textbf{6.90} & 31.80 & 70.15 & \textbf{89.95} \\
\midrule
\multirow{2}{*}{MR40k}    & 1 Epoch & \textbf{1.95} & 13.10 & \textbf{36.55} & \textbf{71.20} & 90.00 \\
                          & 3 Epochs & 1.80 & \textbf{13.20} & 36.30 & 70.40 & \textbf{90.35} \\
\bottomrule
\end{tabular}
\end{minipage}
&
\begin{minipage}{\linewidth}
\centering
\includegraphics[width=\linewidth]{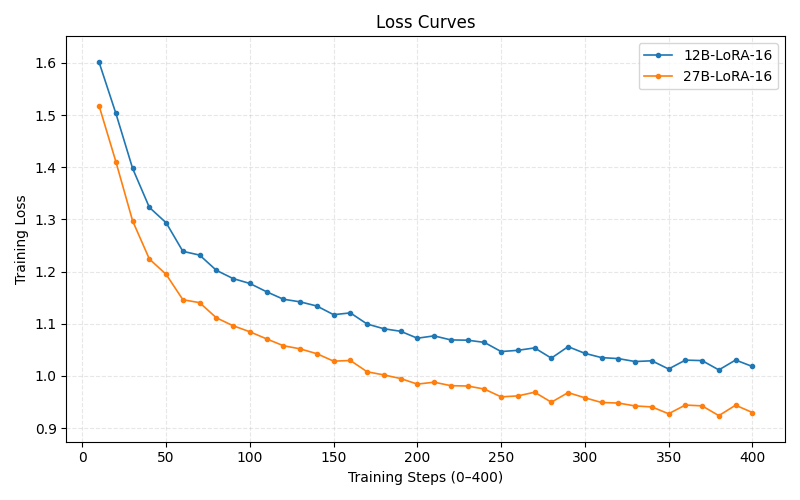}
\captionof{figure}{GeoLocSFT training loss comparison (1 vs. 3 epochs) showing diminishing returns after the first epoch.}
\label{fig:appendix_loss_curve}
\end{minipage}
\end{tabular}
\end{table}

Figure~\ref{fig:appendix_loss_curve} visually demonstrates the training loss progression, supporting the observation that the most significant loss reduction occurs within the first epoch. Table~\ref{tab:appendix_epochs_full_data} presents the full quantitative comparison of geolocation accuracy (Acc@R) on all three benchmarks, confirming that extending training to 3 epochs does not provide substantial benefits over the highly efficient single-epoch fine-tuning. We conclude that 1 epoch of SFT is sufficient and optimal for our GeoLocSFT framework in terms of performance and efficiency.

\subsection{Impact of SFT Data Curation (Diverse vs Biased) - Supporting Table}
\label{subsec:appendix_ablation_data_table}

The main paper (Section~\ref{sec:experiments_ablations}) discusses the impact of SFT data curation, comparing our $\approx$2700 diverse samples against a set of $\approx$1800 samples potentially biased towards certain regions (e.g., Austria). Table~\ref{tab:appendix_data_curation_full_data} provides the quantitative results supporting that discussion.

\begin{table}[!ht]
\centering
\footnotesize
\caption{Ablation Study: Impact of SFT Data Curation ($\approx$2700 Diverse vs $\approx$1800 Biased). Gemma 3 27B-it, 1 epoch SFT (LR=2e-5, Rank=16). Acc@R shown in table.}
\label{tab:appendix_data_curation_full_data}
\setlength{\tabcolsep}{7pt}
\begin{tabular}{@{}llccccc@{}} 
\toprule
\multirow{2}{*}{Benchmark} & \multirow{2}{*}{Data Setting} & \multicolumn{5}{c}{Distance (\% @ km)} \\ 
\cmidrule(lr){3-7} 
& & 1 km & 25 km & 200 km & 750 km & 2500 km \\ 
\midrule
\multirow{2}{*}{Im2GPS3k} & $\approx$1800 Biased & \textbf{9.15} & \textbf{33.10} & \textbf{47.10} & \textbf{65.60} & 82.00 \\
                          & $\approx$2700 Diverse & 8.70 & 32.45 & 47.00 & 65.25 & \textbf{82.15} \\
\midrule
\multirow{2}{*}{GWS15k}   & $\approx$1800 Biased & 0.11 & \textbf{6.55} & 31.80 & 70.75 & 88.50 \\
                          & $\approx$2700 Diverse & \textbf{0.20} & 5.90 & \textbf{34.05} & \textbf{71.00} & \textbf{89.35} \\
\midrule
\multirow{2}{*}{MR40k}    & $\approx$1800 Biased & \textbf{2.25} & \textbf{13.85} & \textbf{37.30} & \textbf{72.40} & \textbf{90.40} \\
                          & $\approx$2700 Diverse & 1.95 & 13.10 & 36.55 & 71.20 & 90.00 \\
\bottomrule
\end{tabular}
\end{table}


\subsection{Hyperparameter Tuning Strategy and Results} 
\label{subsec:appendix_hyperparam_tuning_details} 

Our SFT hyperparameter tuning focused on optimizing learning rate (LR) and LoRA rank ($r$) for single-epoch training, as multi-epoch SFT showed no clear benefit (see Appendix~\ref{subsec:appendix_ablation_epochs}, Table~\ref{tab:appendix_epochs_full_data}, Figure~\ref{fig:appendix_loss_curve}). Our baseline single‑epoch configuration ($LR=2\times10^{-5}$, $r=16$) is shown as Step~3 in Table~\ref{tab:appendix_hyperparam_stacked_results}.

We first hypothesized that a low LoRA rank might limit single-epoch learning. However, a significantly higher rank ($r=256$) with $LR=2 \times 10^{-5}$ (Step 1) did not improve, and sometimes worsened, performance compared to $r=16$ at the same LR. 
Subsequently, we explored various LRs with $r=16$: $LR=3 \times 10^{-5}$ (Step 2) and $LR=1 \times 10^{-5}$ (Step 4). While results varied, none consistently surpassed the Step 3 baseline.

This iterative exploration led to testing $LR=8 \times 10^{-6}$ with a moderate LoRA rank of $r=32$. This combination (Step 5, Table~\ref{tab:appendix_hyperparam_stacked_results}) demonstrated the most robust and superior performance across the GWS15K and MR40k benchmarks.
Thus, $LR=8 \times 10^{-6}$ and LoRA rank $r=32$ (with $\alpha=32$) were selected as optimal for GeoLocSFT, also detailed in Table~\ref{tab:appendix_sft_parameters}.

\begin{table*}[htbp] 
\centering
\small
\setlength{\tabcolsep}{2pt} 
\begin{tabular}{@{}lccllccccc@{}}
\toprule
Step & LR & LoRA Rank ($r$) & Rationale/Observation & Benchmark & 1km & 25km & 200km & 750km & 2500km \\
\midrule
\multirow{2}{*}{1} & \multirow{2}{*}{$2 \times 10^{-5}$} & \multirow{2}{*}{256} & \multirow{2}{*}{High-rank attempt} & 
    GWS15K & 0.10 & 4.80 & 29.30 & 68.25 & 87.55 \\
& & & & MR40k & 1.50 & 12.20 & 34.90 & 69.30 & \textbf{90.35} \\
\cmidrule(l){5-10} 

\multirow{2}{*}{2} & \multirow{2}{*}{$3 \times 10^{-5}$} & \multirow{2}{*}{16} & \multirow{2}{*}{Higher LR, smaller rank} & 
    GWS15K & 0.10 & 6.35 & 32.45 & 70.20 & 88.65 \\
& & & & MR40k & \textbf{2.00} & 12.85 & 35.80 & \textbf{71.30} & 88.95 \\
\cmidrule(l){5-10}

\multirow{2}{*}{3} & \multirow{2}{*}{$2 \times 10^{-5}$} & \multirow{2}{*}{16} & \multirow{2}{*}{Standard LR, smaller rank} & 
    GWS15K & 0.10 & 5.90 & 34.05 & \textbf{71.00} & \textbf{89.35} \\ 
& & & & MR40k & 1.95 & \textbf{13.10} & 36.55 & 71.20 & 90.00 \\
\cmidrule(l){5-10}

\multirow{2}{*}{4} & \multirow{2}{*}{$1 \times 10^{-5}$} & \multirow{2}{*}{16} & \multirow{2}{*}{Lower LR, smaller rank} & 
    GWS15K & 0.05 & \textbf{6.45} & 33.40 & 69.85 & 87.80 \\
& & & & MR40k & 1.35 & 12.90 & \textbf{37.50} & 71.90 & 88.95 \\
\cmidrule(l){5-10}

\multirow{2}{*}{\textbf{5}} & \multirow{2}{*}{\textbf{$8 \times 10^{-6}$}} & \multirow{2}{*}{\textbf{32}} & \multirow{2}{*}{\textbf{Optimal balance}} & 
    GWS15K & \textbf{0.20} & 6.30 & \textbf{33.50} & 69.65 & 87.55\\

& & & & MR40k & 1.70 & 12.75 & 37.50 & 70.85 & 88.95 \\
\bottomrule
\end{tabular}
\caption{Hyperparameter tuning: Acc@R (\%) on GWS15K and MR40k (Ours) for different SFT settings (Steps 1-4) leading to the optimal configuration (Step 5, highlighted).}
\label{tab:appendix_hyperparam_stacked_results}
\end{table*}

\newpage
\section*{NeurIPS Paper Checklist}

\subsection*{1. Claims}
Question: Do the main claims made in the abstract and introduction accurately reflect the paper's contributions and scope?

Answer: {\color{blue}[Yes]}

Justification: Our abstract and introduction clearly outline the three key contributions of our work: the GeoLocSFT framework, exploration of multi-candidate inference strategies, and introduction of the MR40k benchmark for challenging geolocation tasks.

Guidelines:
\begin{itemize}
\item The answer NA means that the abstract and introduction do not include the claims made in the paper.
\item The abstract and/or introduction should clearly state the claims made, including the contributions made in the paper and important assumptions and limitations. A No or NA answer to this question will not be perceived well by the reviewers.
\item The claims made should match theoretical and experimental results, and reflect how much the results can be expected to generalize to other settings.
\item It is fine to include aspirational goals as motivation as long as it is clear that these goals are not attained by the paper.
\end{itemize}

\subsection*{2. Limitations}
Question: Does the paper discuss the limitations of the work performed by the authors?

Answer: {\color{blue}[Yes]}

Justification: In our ablation studies, we acknowledge limitations in our approach, particularly in sampling and aggregation strategies, and identify challenges in prediction confidence assessment that point to areas for future research.

Guidelines:
\begin{itemize}
\item The answer NA means that the paper has no limitation while the answer No means that the paper has limitations, but those are not discussed in the paper.
\item The authors are encouraged to create a separate `Limitations'' section in their paper.
\item The paper should point out any strong assumptions and how robust the results are to violations of these assumptions (e.g., independence assumptions, noiseless settings, model well-specification, asymptotic approximations only holding locally). The authors should reflect on how these assumptions might be violated in practice and what the implications would be.
\item The authors should reflect on the scope of the claims made, e.g., if the approach was only tested on a few datasets or with a few runs. In general, empirical results often depend on implicit assumptions, which should be articulated.
\item The authors should reflect on the factors that influence the performance of the approach. For example, a facial recognition algorithm may perform poorly when image resolution is low or images are taken in low lighting. Or a speech-to-text system might not be used reliably to provide closed captions for online lectures because it fails to handle technical jargon.
\item The authors should discuss the computational efficiency of the proposed algorithms and how they scale with dataset size.
\item If applicable, the authors should discuss possible limitations of their approach to address problems of privacy and fairness.
\item While the authors might fear that complete honesty about limitations might be used by reviewers as grounds for rejection, a worse outcome might be that reviewers discover limitations that aren't acknowledged in the paper. The authors should use their best judgment and recognize that individual actions in favor of transparency play an important role in developing norms that preserve the integrity of the community. Reviewers will be specifically instructed to not penalize honesty concerning limitations.
\end{itemize}

\subsection*{3. Theory Assumptions and Proofs}
Question: For each theoretical result, does the paper provide the full set of assumptions and a complete (and correct) proof?

Answer: {\color{red}[NA]}

Justification: Our paper focuses on empirical methods for visual geolocation and does not include theoretical results requiring formal proofs.

Guidelines:
\begin{itemize}
\item The answer NA means that the paper does not include theoretical results.
\item All the theorems, formulas, and proofs in the paper should be numbered and cross-referenced.
\item All assumptions should be clearly stated or referenced in the statement of any theorems.
\item The proofs can either appear in the main paper or the supplemental material, but if they appear in the supplemental material, the authors are encouraged to provide a short proof sketch to provide intuition.
\item Inversely, any informal proof provided in the core of the paper should be complemented by formal proofs provided in appendix or supplemental material.
\item Theorems and Lemmas that the proof relies upon should be properly referenced.
\end{itemize}

\subsection*{4. Experimental Result Reproducibility}
Question: Does the paper fully disclose all the information needed to reproduce the main experimental results of the paper to the extent that it affects the main claims and/or conclusions of the paper (regardless of whether the code and data are provided or not)?

Answer: {\color{blue}[Yes]}

Justification: We provide details on our model architecture, SFT methodology, dataset descriptions, evaluation metrics, and inference procedures. The appendices include hyperparameters for SFT and information about computational resources used.

Guidelines:
\begin{itemize}
\item The answer NA means that the paper does not include experiments.
\item If the paper includes experiments, a No answer to this question will not be perceived well by the reviewers: Making the paper reproducible is important, regardless of whether the code and data are provided or not.
\item If the contribution is a dataset and/or model, the authors should describe the steps taken to make their results reproducible or verifiable.
\item Depending on the contribution, reproducibility can be accomplished in various ways. For example, if the contribution is a novel architecture, describing the architecture fully might suffice, or if the contribution is a specific model and empirical evaluation, it may be necessary to either make it possible for others to replicate the model with the same dataset, or provide access to the model. In general. releasing code and data is often one good way to accomplish this, but reproducibility can also be provided via detailed instructions for how to replicate the results, access to a hosted model (e.g., in the case of a large language model), releasing of a model checkpoint, or other means that are appropriate to the research performed.
\item While NeurIPS does not require releasing code, the conference does require all submissions to provide some reasonable avenue for reproducibility, which may depend on the nature of the contribution. For example
  (a) If the contribution is primarily a new algorithm, the paper should make it clear how to reproduce that algorithm.
  (b) If the contribution is primarily a new model architecture, the paper should describe the architecture clearly and fully.
  (c) If the contribution is a new model (e.g., a large language model), then there should either be a way to access this model for reproducing the results or a way to reproduce the model (e.g., with an open-source dataset or instructions for how to construct the dataset).
  (d) We recognize that reproducibility may be tricky in some cases, in which case authors are welcome to describe the particular way they provide for reproducibility. In the case of closed-source models, it may be that access to the model is limited in some way (e.g., to registered users), but it should be possible for other researchers to have some path to reproducing or verifying the results.
\end{itemize}

\subsection*{5. Open access to data and code}
Question: Does the paper provide open access to the data and code, with sufficient instructions to faithfully reproduce the main experimental results, as described in supplemental material?

Answer: {\color{blue}[Yes]}

Justification: We will publicly release our code and the MR40k benchmark dataset upon publication. Our appendices provide detailed instructions regarding implementation, hyperparameters, and experimental setup to facilitate reproduction.

Guidelines:
\begin{itemize}
\item The answer NA means that paper does not include experiments requiring code.
\item Please see the NeurIPS code and data submission \href{https://nips.cc/public/guides/CodeSubmissionPolicy}{guidelines} for more details.
\item While we encourage the release of code and data, we understand that this might not be possible, so "No" is an acceptable answer. Papers cannot be rejected simply for not including code, unless this is central to the contribution (e.g., for a new open-source benchmark).
\item The instructions should contain the exact command and environment needed to run to reproduce the results. See the NeurIPS code and data submission \href{https://nips.cc/public/guides/CodeSubmissionPolicy}{guidelines} for more details.
\item The authors should provide instructions on data access and preparation, including how to access the raw data, preprocessed data, intermediate data, and generated data, etc.
\item The authors should provide scripts to reproduce all experimental results for the new proposed method and baselines. If only a subset of experiments are reproducible, they should state which ones are omitted from the script and why.
\item At submission time, to preserve anonymity, the authors should release anonymized versions (if applicable).
\item Providing as much information as possible in supplemental material (appended to the paper) is recommended, but including URLs to data and code is permitted.
\end{itemize}

\subsection*{6. Experimental Setting/Details}
Question: Does the paper specify all the training and test details (e.g., data splits, hyperparameters, how they were chosen, type of optimizer, etc.) necessary to understand the results?

Answer: {\color{blue}[Yes]}

Justification: Our experimental section and appendices provide comprehensive information on dataset partitioning, evaluation metrics, implementation details, SFT parameters, model architecture, and inference strategies.

Guidelines:
\begin{itemize}
\item The answer NA means that the paper does not include experiments.
\item The experimental setting should be presented in the core of the paper to a level of detail that is necessary to appreciate the results and make sense of them.
\item The full details can be provided either with the code, in appendix, or as supplemental material.
\end{itemize}

\subsection*{7. Experiment Statistical Significance}
Question: Does the paper report error bars suitably and correctly defined or other appropriate information about the statistical significance of the experiments?

Answer: {\color{gray}[No]}

Justification: While we do not report error bars, we perform comprehensive ablation studies across multiple benchmarks to validate the robustness of our findings, testing across different configurations to ensure consistent improvements.

Guidelines:
\begin{itemize}
\item The answer NA means that the paper does not include experiments.
\item The authors should answer "Yes" if the results are accompanied by error bars, confidence intervals, or statistical significance tests, at least for the experiments that support the main claims of the paper.
\item The factors of variability that the error bars are capturing should be clearly stated (for example, train/test split, initialization, random drawing of some parameter, or overall run with given experimental conditions).
\item The method for calculating the error bars should be explained (closed form formula, call to a library function, bootstrap, etc.)
\item The assumptions made should be given (e.g., Normally distributed errors).
\item It should be clear whether the error bar is the standard deviation or the standard error of the mean.
\item It is OK to report 1-sigma error bars, but one should state it. The authors should preferably report a 2-sigma error bar than state that they have a 96\% CI, if the hypothesis of Normality of errors is not verified.
\item For asymmetric distributions, the authors should be careful not to show in tables or figures symmetric error bars that would yield results that are out of range (e.g. negative error rates).
\item If error bars are reported in tables or plots, The authors should explain in the text how they were calculated and reference the corresponding figures or tables in the text.
\end{itemize}

\subsection*{8. Experiments Compute Resources}
Question: For each experiment, does the paper provide sufficient information on the computer resources (type of compute workers, memory, time of execution) needed to reproduce the experiments?

Answer: {\color{blue}[Yes]}

Justification: We specify that our SFT was performed on a cluster of 8 NVIDIA A100 GPUs, with the 1-epoch training process completing in approximately 50 minutes. We also report model sizes and memory requirements.

Guidelines:
\begin{itemize}
\item The answer NA means that the paper does not include experiments.
\item The paper should indicate the type of compute workers CPU or GPU, internal cluster, or cloud provider, including relevant memory and storage.
\item The paper should provide the amount of compute required for each of the individual experimental runs as well as estimate the total compute.
\item The paper should disclose whether the full research project required more compute than the experiments reported in the paper (e.g., preliminary or failed experiments that didn't make it into the paper).
\end{itemize}

\subsection*{9. Code Of Ethics}
Question: Does the research conducted in the paper conform, in every respect, with the NeurIPS Code of Ethics https://neurips.cc/public/EthicsGuidelines?

Answer: {\color{blue}[Yes]}

Justification: Our research adheres to the NeurIPS Code of Ethics. We use publicly available datasets, acknowledge relevant prior work, and are transparent about our methodologies. We also explicitly acknowledge ethical considerations related to geolocation technologies.

Guidelines:
\begin{itemize}
\item The answer NA means that the authors have not reviewed the NeurIPS Code of Ethics.
\item If the authors answer No, they should explain the special circumstances that require a deviation from the Code of Ethics.
\item The authors should make sure to preserve anonymity (e.g., if there is a special consideration due to laws or regulations in their jurisdiction).
\end{itemize}

\subsection*{10. Broader Impacts}
Question: Does the paper discuss both potential positive societal impacts and negative societal impacts of the work performed?

Answer: {\color{blue}[Yes]}

Justification: We acknowledge that visual geolocation has positive applications such as photo organization and journalistic verification, while also explicitly recognizing ethical considerations and the need for responsible development to mitigate potential misuse.

Guidelines:
\begin{itemize}
\item The answer NA means that there is no societal impact of the work performed.
\item If the authors answer NA or No, they should explain why their work has no societal impact or why the paper does not address societal impact.
\item Examples of negative societal impacts include potential malicious or unintended uses (e.g., disinformation, generating fake profiles, surveillance), fairness considerations (e.g., deployment of technologies that could make decisions that unfairly impact specific groups), privacy considerations, and security considerations.
\item The conference expects that many papers will be foundational research and not tied to particular applications, let alone deployments. However, if there is a direct path to any negative applications, the authors should point it out. For example, it is legitimate to point out that an improvement in the quality of generative models could be used to generate deepfakes for disinformation. On the other hand, it is not needed to point out that a generic algorithm for optimizing neural networks could enable people to train models that generate Deepfakes faster.
\item The authors should consider possible harms that could arise when the technology is being used as intended and functioning correctly, harms that could arise when the technology is being used as intended but gives incorrect results, and harms following from (intentional or unintentional) misuse of the technology.
\item If there are negative societal impacts, the authors could also discuss possible mitigation strategies (e.g., gated release of models, providing defenses in addition to attacks, mechanisms for monitoring misuse, mechanisms to monitor how a system learns from feedback over time, improving the efficiency and accessibility of ML).
\end{itemize}

\subsection*{11. Safeguards}
Question: Does the paper describe safeguards that have been put in place for responsible release of data or models that have a high risk for misuse (e.g., pretrained language models, image generators, or scraped datasets)?

Answer: {\color{red}[NA]}

Justification: Our research focuses on geolocation methods that don't present high risks for misuse in the same way as generative models or datasets containing sensitive information might.

Guidelines:
\begin{itemize}
\item The answer NA means that the paper poses no such risks.
\item Released models that have a high risk for misuse or dual-use should be released with necessary safeguards to allow for controlled use of the model, for example by requiring that users adhere to usage guidelines or restrictions to access the model or implementing safety filters.
\item Datasets that have been scraped from the Internet could pose safety risks. The authors should describe how they avoided releasing unsafe images.
\item We recognize that providing effective safeguards is challenging, and many papers do not require this, but we encourage authors to take this into account and make a best faith effort.
\end{itemize}

\subsection*{12. Licenses for existing assets}
Question: Are the creators or original owners of assets (e.g., code, data, models), used in the paper, properly credited and are the license and terms of use explicitly mentioned and properly respected?

Answer: {\color{blue}[Yes]}

Justification: We properly credit and cite all existing assets used in our research, including the base models, datasets, and prior methodologies that our work builds upon.

Guidelines:
\begin{itemize}
\item The answer NA means that the paper does not use existing assets.
\item The authors should cite the original paper that produced the code package or dataset.
\item The authors should state which version of the asset is used and, if possible, include a URL.
\item The name of the license (e.g., CC-BY 4.0) should be included for each asset.
\item For scraped data from a particular source (e.g., website), the copyright and terms of service of that source should be provided.
\item If assets are released, the license, copyright information, and terms of use in the package should be provided. For popular datasets, paperswithcode.com/datasets has curated licenses for some datasets. Their licensing guide can help determine the license of a dataset.
\item For existing datasets that are re-packaged, both the original license and the license of the derived asset (if it has changed) should be provided.
\item If this information is not available online, the authors are encouraged to reach out to the asset's creators.
\end{itemize}

\subsection*{13. New Assets}
Question: Are new assets introduced in the paper well documented and is the documentation provided alongside the assets?

Answer: {\color{blue}[Yes]}

Justification: Our MR40k benchmark dataset is thoroughly documented, including details on data collection, filtering criteria, geographic distribution, and intended use. This documentation will be released alongside the dataset.

Guidelines:
\begin{itemize}
\item The answer NA means that the paper does not release new assets.
\item Researchers should communicate the details of the dataset/code/model as part of their submissions via structured templates. This includes details about training, license, limitations, etc.
\item The paper should discuss whether and how consent was obtained from people whose asset is used.
\item At submission time, remember to anonymize your assets (if applicable). You can either create an anonymized URL or include an anonymized zip file.
\end{itemize}

\subsection*{14. Crowdsourcing and Research with Human Subjects}
Question: For crowdsourcing experiments and research with human subjects, does the paper include the full text of instructions given to participants and screenshots, if applicable, as well as details about compensation (if any)?

Answer: {\color{red}[NA]}

Justification: Our research does not involve crowdsourcing or human subjects.

Guidelines:
\begin{itemize}
\item The answer NA means that the paper does not involve crowdsourcing nor research with human subjects.
\item Including this information in the supplemental material is fine, but if the main contribution of the paper involves human subjects, then as much detail as possible should be included in the main paper.
\item According to the NeurIPS Code of Ethics, workers involved in data collection, curation, or other labor should be paid at least the minimum wage in the country of the data collector.
\end{itemize}

\subsection*{15. Institutional Review Board (IRB) Approvals}
Question: Does the paper describe potential risks incurred by study participants, whether such risks were disclosed to the subjects, and whether Institutional Review Board (IRB) approvals (or an equivalent approval/review based on the requirements of your country or institution) were obtained?

Answer: {\color{red}[NA]}

Justification: Our research does not involve human subjects, so IRB approval was not required.

Guidelines:
\begin{itemize}
\item The answer NA means that the paper does not involve crowdsourcing nor research with human subjects.
\item Depending on the country in which research is conducted, IRB approval (or equivalent) may be required for any human subjects research. If you obtained IRB approval, you should clearly state this in the paper.
\item We recognize that the procedures for this may vary significantly between institutions and locations, and we expect authors to adhere to the NeurIPS Code of Ethics and the guidelines for their institution.
\item For initial submissions, do not include any information that would break anonymity (if applicable), such as the institution conducting the review.
\end{itemize} 

\end{document}